\documentclass{article}

\usepackage{arxiv}
\usepackage{authblk}
\usepackage[utf8]{inputenc} 
\usepackage[T1]{fontenc}    
\usepackage{url}            
\usepackage{booktabs}       
\usepackage{nicefrac}       
\usepackage{microtype}      
\usepackage{natbib} 
\usepackage{amsmath}
\usepackage{amssymb}
\usepackage{amsfonts}
\usepackage{amsthm}
\usepackage{hyperref}
\usepackage{graphicx}
\usepackage{algorithm}
\usepackage{algpseudocode}
\usepackage{subcaption}
\usepackage{enumitem}
\usepackage[normalem]{ulem}
\usepackage{comment}

\newcommand{\E}{\mathbb{E}} 
\newcommand{\R}{\mathbb{R}} 
\newtheorem{theorem}{Theorem}[section]
\newtheorem{proposition}[theorem]{Proposition}

\theoremstyle{definition}

\title{Stochastic Gradient Descent in the Saddle-to-Saddle Regime of Deep Linear Networks}

\author[1]{Guillaume Corlouer\textsuperscript{\hspace{0.08em}*}}
\author[2]{Avi Semler}
\author[3]{Alexander Strang}
\author[4]{Alexander Gietelink Oldenziel}

\affil[1]{Moirai}
\affil[2]{University of Oxford}
\affil[3]{University of California, Berkeley}
\affil[4]{Iliad}

\begin{document}
\maketitle
\begin{abstract}
Deep linear networks (DLNs) are used as an analytically tractable model of the training dynamics of deep neural networks.
While gradient descent in DLNs is known to exhibit saddle-to-saddle dynamics, the impact of stochastic gradient descent (SGD) noise on this regime remains poorly understood.
We investigate the dynamics of SGD during training of DLNs in the saddle-to-saddle regime. We model the training dynamics as stochastic Langevin dynamics with anisotropic, state-dependent noise. Under the assumption of aligned and balanced weights, we derive an exact decomposition of the dynamics into a system of one-dimensional per-mode stochastic differential equations. This establishes that the maximal diffusion along a mode precedes the corresponding feature being completely learned. We also derive the stationary distribution of SGD for each mode: in the absence of label noise, its marginal distribution along specific features coincides with the stationary distribution of gradient flow, while in the presence of label noise it approximates a Boltzmann distribution. Finally, we confirm experimentally that the theoretical results hold qualitatively even without aligned or balanced weights. These results establish that SGD noise encodes information about the progression of feature learning but does not fundamentally alter the saddle-to-saddle dynamics.

\end{abstract}
\tableofcontents
\newpage

\section{Introduction}
Stochastic gradient descent (SGD) and its variants are optimization algorithms widely used to train deep neural networks. Classical statistical learning theory struggles to account for the observed ability of deep neural networks to generalize beyond their training dataset \citep{Zhang2017Rethinking}. Therefore, it is proposed that there must be an implicit bias in the learning algorithm that emerges from the interaction of data, architecture and optimizer. Understanding this bias may be important for AI alignment and safety \citep{lehalleur2025you, DBLP:journals/tmlr/AnwarSRPTHLJCSE24}.

SGD differs from gradient descent by computing the update direction based on a randomly sampled subset of the data that varies between updates, rather than using a fixed dataset for all of training. One can distinguish the contribution of two terms during SGD. The first term is the gradient, and the second term is the stochasticity, which comes from the randomness in approximating the gradient. It is unclear whether the stochasticity is crucial in shaping the implicit bias, or if it is merely computationally convenient: for example, \cite{paquette2022implicitregularizationimplicitconditioning} shows that for high-dimensional convex optimization the noise is not important, and \cite{vyas2024implicitbiasinsignificancesgd} provides empirical evidence that the regimes under which stochasticity is not relevant for generalization. However, \cite{pesme2021implicit} shows that for diagonal linear networks it promotes sparsity.

Deep linear networks (DLNs) are a simple class of neural networks, consisting only of matrix multiplication operations. Despite only expressing linear functions, their training dynamics are nonlinear and exhibit many of the interesting phenomena that occur in architectures with nonlinearities \citep{nam2025position}. We choose DLNs as the setting for this work because it makes precise mathematical analysis tractable. 

A key result in this literature is that gradient flow on DLNs, under small initialization, proceeds through a saddle-to-saddle regime. In this regime, the network traverses a sequence of saddle points, learning the singular values of the target ("teacher") matrix in decreasing order of magnitude \citep{jacot2021saddle}. This stage-wise, time-scale-separated dynamics has been characterized exactly for gradient flow \citep{saxe2013exact}, and extensions to stochastic gradient flow in diagonal and rank-one linear networks have been explored \citep{pesme2021implicit, lyu2023implicit}. However, the impact of using a continuous model of SGD rather than gradient flow has not been analytically characterized for fully connected DLNs.

\subsection{Contributions}

We study the training dynamics of SGD modeled as a stochastic differential equation (SDE) on deep linear networks. More specifically, assuming that the weights are balanced and aligned during training, we model SGD using its continuous limit as an Itô SDE and decompose it into a system of one-dimensional SDEs. We focus on the saddle-to-saddle regime \cite{jacot2021saddle}, during which the singular values of a teacher matrix are learned in parallel and at different time scales. This extends the gradient-flow analysis of \citet{saxe2013exact} to a stochastic setting. Our main contributions are:

\begin{enumerate}
\item \textbf{Exact SGD noise covariance.} We derive a closed-form expression for the gradient noise covariance matrix of SGD in DLNs, both with and without label noise. This expression is state-dependent and anisotropic, and it decomposes cleanly into a data-mismatch term and a label-noise term.
\item \textbf{Modewise diffusion predicts feature learning.} Under the balanced and aligned assumptions, we show that the diffusion coefficient along a given mode peaks \emph{before} that mode is fully learned, then decays to zero once the mode has been fully learned (shown in \autoref{fig:summary}). This establishes that SGD noise carries information about the progression of feature learning.

\item \textbf{Stationary modewise distributions.} We characterize the stationary distribution of the modewise SDE via detailed balance. In the absence of label noise, the stationary distribution collapses to a Dirac mass, thus matching the gradient flow solution. With label noise, it is approximately Boltzmann.

\item \textbf{State-dependent noise is a more accurate model.} We also find that a continuous model of SGD with state-dependent noise is a more accurate model of SGD than the isotropic homogeneous noise (Langevin), which is commonly assumed in the literature during the feature learning regime and for the end-of-training distribution.

\end{enumerate}
Together, these results show that SGD noise encodes information about the stage of learning, but does not qualitatively alter the saddle-to-saddle structure: modes are still learned in order of decreasing singular value magnitude, with SGD primarily affecting the timescale of each transition. We also verify experimentally that the qualitative predictions hold even when the balanced and aligned assumptions are relaxed.

\subsection{Related work}
\subsubsection{Implicit biases of SGD noise}
 Some previous work argues that SGD noise matters for generalization. More specifically, gradient noise induces an implicit bias in SGD that attracts dynamics towards invariant sets of the parameter space corresponding to simpler subnetworks. This manifests as a noise-induced drift that pulls parameters toward zero, making neurons vanish or become redundant \citep{chen2023stochastic}. In diagonal linear networks, stochastic gradient flow has an implicit bias toward sparser solutions that is not present in gradient flow, suggesting that stochasticity matters for generalization \citep{pesme2021implicit}. In non-linear deep neural networks, during loss stabilization, the combination of gradient and noise also induces a bias towards sparser solutions when the learning rate is large \citep{andriushchenko2023sgd}. Furthermore, some implicit biases of SGD can be made explicit by showing that SGD achieves the same performance as GD with an explicit regularization term that penalizes large batch-gradient updates \citep{geiping2021stochastic}.  In linear networks, stochastic gradient flow appears to be less dependent on initialization than gradient flow and induces an additional bias towards simpler solutions beyond the simplicity bias of gradient flow \citep{varre2024sgd}.  

Other work has investigated the structure of SGD noise, which is state-dependent (i.e., it differs between points in parameter space) and anisotropic (the distribution of the noise is not rotationally invariant). The structure of the noise could matter for generalization and understanding the training dynamics of SGD. For example, in deep linear networks, SGD structured noise does not allow jumps from lower-rank to higher-rank weight matrices, while the noise from a Langevin process has a non-zero probability of jumping back to higher rank solutions \citep{wang2023implicit}. Furthermore, the structured noise of SGD is sensitive to geometry and can induce an implicit bias towards flatter minima \citep{xie2020diffusion}. In particular, critical points of the loss landscapes of deep neural networks are typically highly degenerate \citep{sagun2017eigenvalues}, and SGD noise is sensitive to these degeneracies by slowing down along degenerate directions. This slowing effect is not present with Langevin dynamics \citep{corlouerSGDsticky}.  In addition to being structured, SGD noise can also be autocorrelated (colored noise). In a dynamical mean field theoretic model, SGD noise can converge to a non-equilibrium steady-state solution, where noisier regimes are associated with solutions that are more robust due to having wider decision boundaries \citep{mignacco2022effective}. 

However, other work suggests that the implicit biases arising from stochasticity do not matter in some regimes.  In particular, the \textit{Golden Path} hypothesis states that the population loss of gradient descent is upper bounded by the population loss of SGD for a given trajectory with fixed initialization in the online learning regime in which new batches are sampled at each time step. Empirical evidence for this Golden Path hypothesis has been found by showing that switching from high noise to low noise during training leads to convergence to the same solution as when using only low noise, for convolutional neural networks and transformers \citep{vyas2024implicitbiasinsignificancesgd}. \cite{paquette2022implicitregularizationimplicitconditioning} prove that a Golden Path hypothesis holds for convex quadratic loss landscapes in high dimensions, using a novel continuous model of SGD \citep{paquette2024homogenization, mignacco2022effective}.

\subsubsection{Regimes of learning in Deep Linear Networks}
Deep linear networks (DLNs) serve as an analytically tractable toy model that can shed light on the training dynamics of non-linear deep neural networks. DLN training has a rich non-linear dynamics and a non-convex high-dimensional loss landscape despite the expressivity of DLNs being limited to linear functions \citep{nam2025position}. A particularly important result is the exact solution of gradient flow on DLNs \citep{saxe2013exact} during the feature learning regime. In this regime, the training dynamics undergoes a separation of time-scales in which a DLN learns the singular values of the teacher matrix in decreasing order of size. This feature learning regime corresponds to a saddle-to-saddle dynamics in which gradient flow traverses the loss landscape through a series of saddle points in which the loss is stabilized until the flow can escape to the next saddle, which increases the rank of the solution by one \citep{jacot2021saddle}.

This regime of saddle-to-saddle dynamics (also called ``rich'' regime) contrasts with the ``lazy'' regime in which the neural tangent kernel (NTK) at initialization––a linear operator––determines the time evolution of the network's function \citep{jacot2018neural}. The regime of training is determined by hyperparameters such as the variance of the parameters at initialization or the width of the neural network \citep{domine2024lazy}. Transitions between regimes are possible: for example, the \emph{grokking} phenomenon has been hypothesized to be a transition from a lazy to a rich regime \citep{kumar2310grokking}. 

In the limit of large depth, width, and amount of data (with constant ratios between these quantities), the generalization error of gradient flow has been characterized under different parametrizations with dynamical mean field theory, which enables theoretical predictions about gains from increased width and scaling laws in the training curve for some structured data \citep{bordelon2025deep}.

Importantly, the training dynamics of gradient flow in DLNs is well understood \citep{advani2020high}, and extensions to stochastic gradient flow have been explored in diagonal linear networks \citep{pesme2021implicit} and rank-one linear networks \citep{lyu2023implicit}. Additionally, SGD noise anisotropy causes the weights' fluctuations during training to be inversely proportional to the flatness of the loss landscape in two-layer DLNs \citep{gross2024weight}. However, the training dynamics of SGD in fully connected DLNs remain to be understood in the rich (saddle-to-saddle) regime.

\subsubsection{Steady-state distribution of SGD}
Another facet of the training dynamics is understanding the convergence properties of SGD, and specifically its end-of-training distribution. Under the assumptions that SGD is well approximated by Langevin dynamics, i.e. that SGD noise is white noise (Gaussian, with constant isotropic covariance) and that the loss is non-degenerate, then SGD approximates Bayesian inference and its limiting distribution is a Boltzmann distribution \citep{mandt2017stochastic, welling2011bayesian}.  However, SGD noise is anisotropic and state-dependent, and the loss of neural networks is highly degenerate, which induces differences from the Bayesian approximation. For example, unlike a Bayesian learner, SGD can get stuck along a degenerate direction of a critical submanifold of the loss landscape \citep{corlouerSGDsticky}. Additionally, degeneracies and noise anisotropy can induce a non-equilibrium steady-state distribution with circular currents where the weights oscillate around critical points \citep{chaudhari2018stochastic, kunin2023limiting}.  The end of distribution of SGD can be better understood if we model SGD as optimizing a competition between an energy and an entropy term corresponding to a Helmholtz Free Energy functional \citep{sadrtdinov2025sgd,chaudhari2018stochastic}.

Another intriguing phenomenon is the anomalous diffusion of SGD. Specifically, we can observe sub-diffusive behavior of SGD where the mean square displacement of the weights is slower than what would be expected under Brownian motion. At the level of the distribution of SGD trajectories, this can be modeled by a time-fractional Fokker-Planck equation \citep{hennick2025almost}.

\begin{figure}
    \centering
    \begin{subfigure}{0.45\linewidth}
        \includegraphics[width=\linewidth, height=5cm, keepaspectratio]{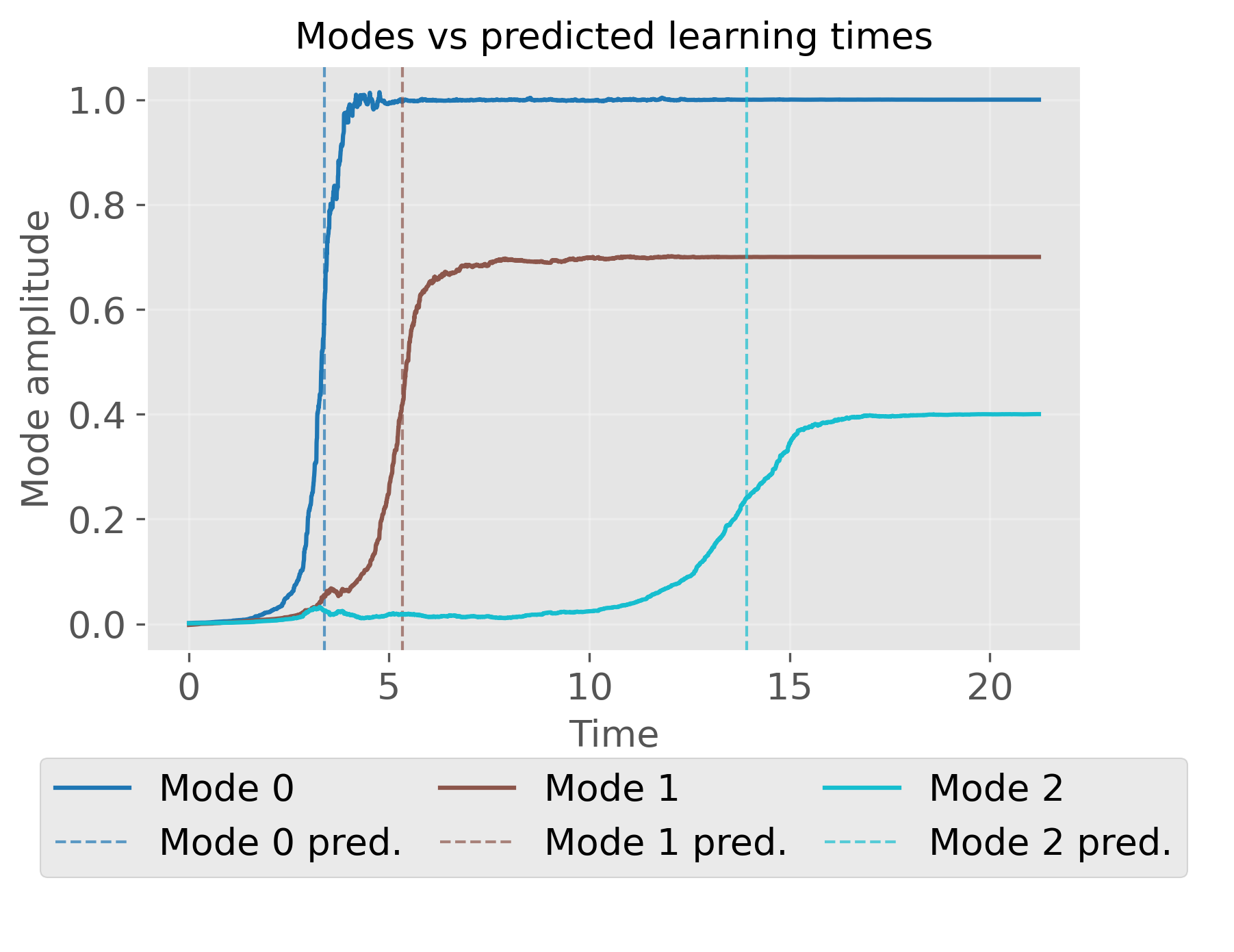}
        \caption{Modes being learned}
    \end{subfigure}
    \begin{subfigure}{0.45\linewidth}
        \includegraphics[width=\linewidth, height=5cm, keepaspectratio]{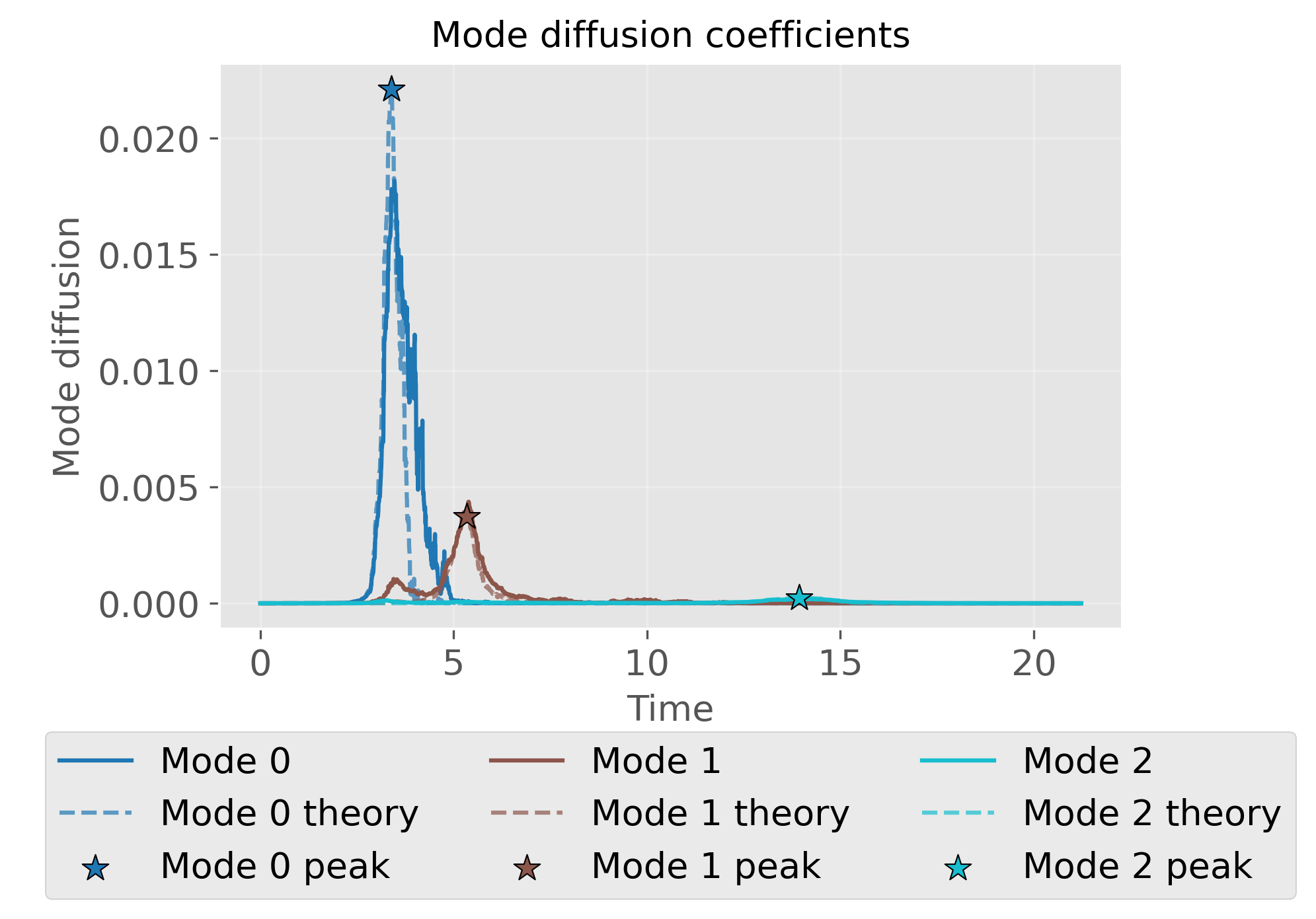}
        \caption{Diffusion along modes}
    \end{subfigure}
    \caption{\textbf{Predicting when modes are learned} using the predicted time of maximum diffusion, in a 4-layer linear network trained with SGD. (a) The modes are learned in order of magnitude, with the time of learning being predicted by the time of maximum diffusion. (b) The diffusion along a mode peaks while a mode is being learned, and our theoretical prediction (see \autoref{eq:diffusion}) matches what is observed. The vertical lines in (a) correspond to the peaks (marked with $\bigstar$ in (b)) of the theoretical prediction for the diffusion (see \autoref{eq:wstar}).}
    \label{fig:summary}
\end{figure}

\section{Preliminaries }
\subsection{Online SGD update}
\label{subsec:sgd-update}
Let $\mathcal{X}\subset\mathbb{R}^{d_0}$ be a set of inputs, and $\mathcal{Y}\subset\mathbb{R}^{d_L}$ be a set of possible outputs. We consider a joint distribution $p_{X, Y}$ over $\mathcal{X} \times \mathcal{Y}$, with marginal  $p_X$ and conditional output distribution $p_{Y \mid X}$. We write $X \sim p_X$ for a random input and 
$Y \mid X \sim p_{Y \mid X}(\,\cdot \mid X)$ for the associated output.

 A deep neural network architecture is a function\footnote{More specifically a composition of linear and non-linear functions, loosely abstracting the function of biological neurons} $f:\Theta\times\mathcal{X}\to\mathcal{Y}$ with parameters $\theta\in\Theta\subseteq\mathbb{R}^d$ which can be trained to learn the expected output $\mathbb{E}[Y|X=x]$ given an input $x$ by minimizing the mean-squared error (MSE) over the data distribution:
\[
L(\theta)=\mathbb{E}_{X,Y}[ \ell(\theta;Y,X)],\qquad
\ell(\theta;y,x)=\tfrac12\|y-f(\theta;x)\|^2.
\]
Because this is often intractable to minimize directly, we instead calculate the empirical batch loss and its gradient on a finite batch $B=\{(x_i, y_i)\}_{i=1}^b$, sampled independently from the distribution: 
\[
L_B(\theta)=\frac{1}{b}\sum_{i=1}^b \ell(\theta;y_i,x_i).
\]
In online\footnote{This contrasts with \emph{offline} SGD, where a finite dataset is sampled from the distribution, and then all batches are sampled from the finite dataset} SGD, the loss is minimized by initializing the neural network parameters as $\theta_0$, and then repeatedly sampling batches and updating the parameters using the empirical batch gradient $g_B(\theta):=\nabla L_B(\theta)$.
Given a batch $B_k\subset(\mathcal{X}\times \mathcal{Y})^b$ of size $b$, the discrete-time SGD update with \textit{learning rate} $\eta_k>0$ is given by:
\[
\theta_{k+1}=\theta_k-\eta_k\,g_{B_k}(\theta_k).
\]
Observe that the batch gradient is an unbiased estimator of the population gradient: $\mathbb{E}_B[g_B(\theta)]=\nabla L(\theta)$.

\subsection{Continuous limit of SGD (constant step size)}
\label{subsec:sgd-to-sde}
Define the one-sample gradient noise $\xi(\theta;X, Y):=g_{(X,Y)}(\theta)-g(\theta)$, a random variable denoting the difference between the batch gradient using $(X,Y)$, and the population gradient $g(\theta):=\nabla L(\theta)$. Its covariance matrix is
\[
\Sigma(\theta):=\mathbb{E}_{X, Y\sim p_{X, Y}}\big[\xi(\theta;X,Y)\,\xi(\theta;X,Y)^\top\big].
\]
For batches of size $b$, the batch-gradient covariance satisfies.
\[
\Sigma_b(\theta)=\frac{1}{b}\Sigma(\theta),
\]
and we see that the batch-gradient noise covariance $\Sigma_b$ is proportional to the one-sample gradient noise covariance. Under the usual martingale functional CLT assumptions (uniform convergence of conditional quadratic variation and Lindeberg condition; see Appendix~\ref{app:sgd-to-sde}), the piecewise-constant interpolation of $\{\theta_k\}$ with constant $\eta$ converges (as $\eta\to 0$) to the Itô SDE
\begin{equation}
\label{eq:sgd-sde}
d\theta_t=-\,g(\theta_t)\,dt+\sqrt{\eta\,\Sigma_b(\theta_t)}\,dW_t,
\end{equation}
with $W_t$ a  Wiener process. Equation~\eqref{eq:sgd-sde} is the continuous-time model of SGD that we will use throughout, and will refer to  as \emph{anisotropic Langevin dynamics}.

\subsection{Deep linear networks (DLNs) and mode dynamics during gradient flow}
\label{subsec:dln-setup}
Let $x\in\mathbb{R}^{d_0}$ and $W_l\in\mathbb{R}^{d_l \times d_{l-1}}$. A depth-$L$ DLN is defined by the following linear input-output map:
\[
f(x)=W_L W_{L-1}\cdots W_1 x = Wx\in \mathbb{R}^{d_L}
\]
Data is generated by a teacher $M\in\mathbb{R}^{d_L\times d_0}$ via $Y=MX$ with whitened Gaussian inputs $X$ such that $\mathbb{E}[XX^\top]=I_{d_0}$. We will also sometimes consider some label noise in addition to the teacher matrix, i.e., $Y=MX +\xi_q$ where $\xi_q \sim N(0,\sigma_qI)$. Let the SVD be $M=USV^\top$, with left and right singular values $(u_\alpha,v_\alpha)$ associated to the singular value $s_\alpha$.
A standard approach in the theory of deep linear networks is to decompose the training dynamics onto \textit{modes}, i.e., onto a particular left and right singular value of the target function. Define the \emph{mode} and \emph{cross-mode amplitude} of the student:
\[
w_\alpha:=u_{\alpha}^{\top}\!\big(W_L\cdots W_1\big)v_\alpha,\qquad
w_{\alpha\beta}:=u_{\alpha}^{\top}\!\big(W_L\cdots W_1\big)v_\beta.
\]
Intuitively, the mode amplitude measures the extent to which the network function has learned a singular value of the teacher function. For example, when a mode has been fully learned, we have $w_{\alpha} = s_{\alpha}$. 
Under balanced initialization ($W_{\ell + 1}^{\top}W_{\ell + 1}=W_{\ell}W_{\ell}^{\top}$), no label noise\footnote{See \citep{advani2020high} for the case with label noise} and orthogonality of distinct modes (i.e. all cross modes are zero), the gradient-flow (GF) dynamics (continuous-time limit of GD) on a depth-2 linear network decouples along modes (see \cite{saxe2013exact} for more details):
\begin{equation}
\label{eq:mode-gf}
\dot w_\alpha \;=\; 2\,(s_\alpha-w_\alpha)\,w_\alpha
\end{equation}
Despite the linearity of DLNs, the latter equation is non-linear in the mode amplitudes. The solution of \eqref{eq:mode-gf} is logistic:
\begin{equation}
\label{eq:mode-logistic}
w_\alpha(t)
=\frac{s_\alpha}{1+\left(\frac{s_\alpha}{w_\alpha(0)}-1\right)e^{-2s_\alpha t}},
\end{equation}
showing \emph{stagewise learning}: larger $s_\alpha$ modes rise earlier and faster (characteristic timescale $\sim 1/s_\alpha$). The training dynamics are nonlinear and, generically, exhibit saddle-to-saddle transients before reaching minimizers; non-strict saddles are prevalent in DLNs and also arise in nonlinear DNNs.

\begin{figure}
    \centering
    \begin{subfigure}{0.49\textwidth}
        \includegraphics[width=\linewidth]{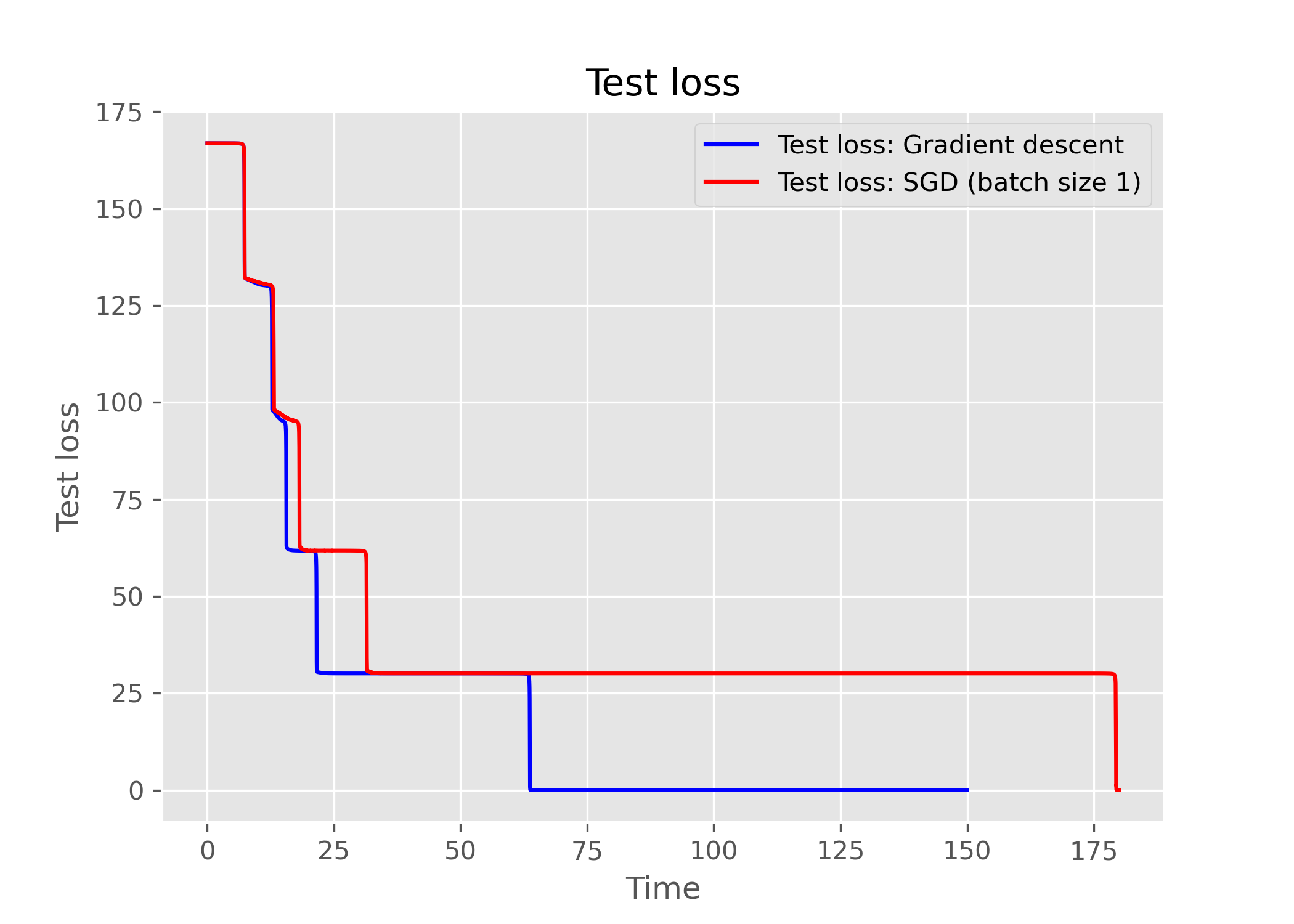}
        \caption{Discrete optimizers}
    \end{subfigure}
    \begin{subfigure}{0.49\textwidth}
        \includegraphics[width=\linewidth]{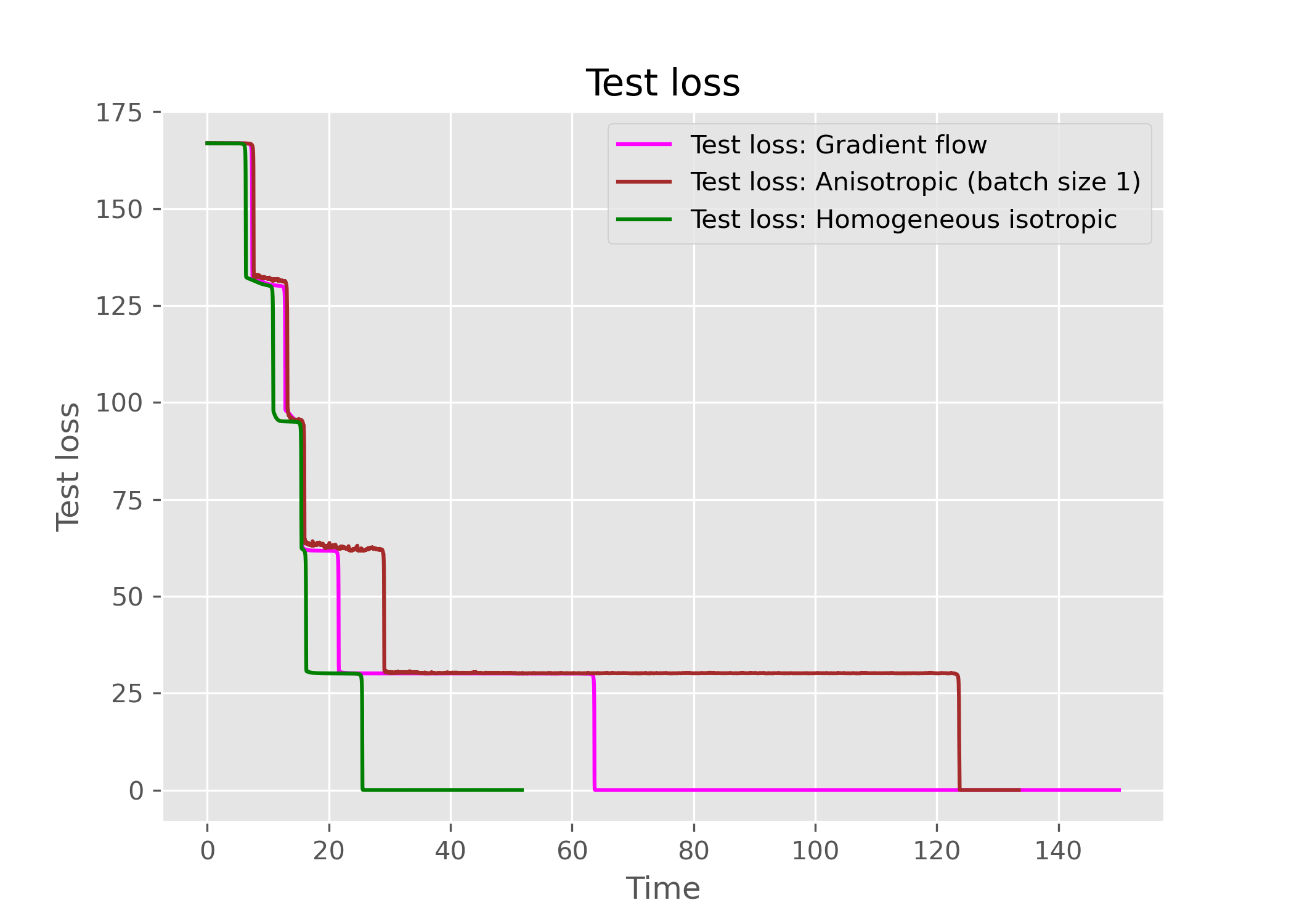}
        \caption{Numerical simulation of continuous limits}
    \end{subfigure}
    \begin{subfigure}{0.49\textwidth}
        \includegraphics[width=\linewidth]{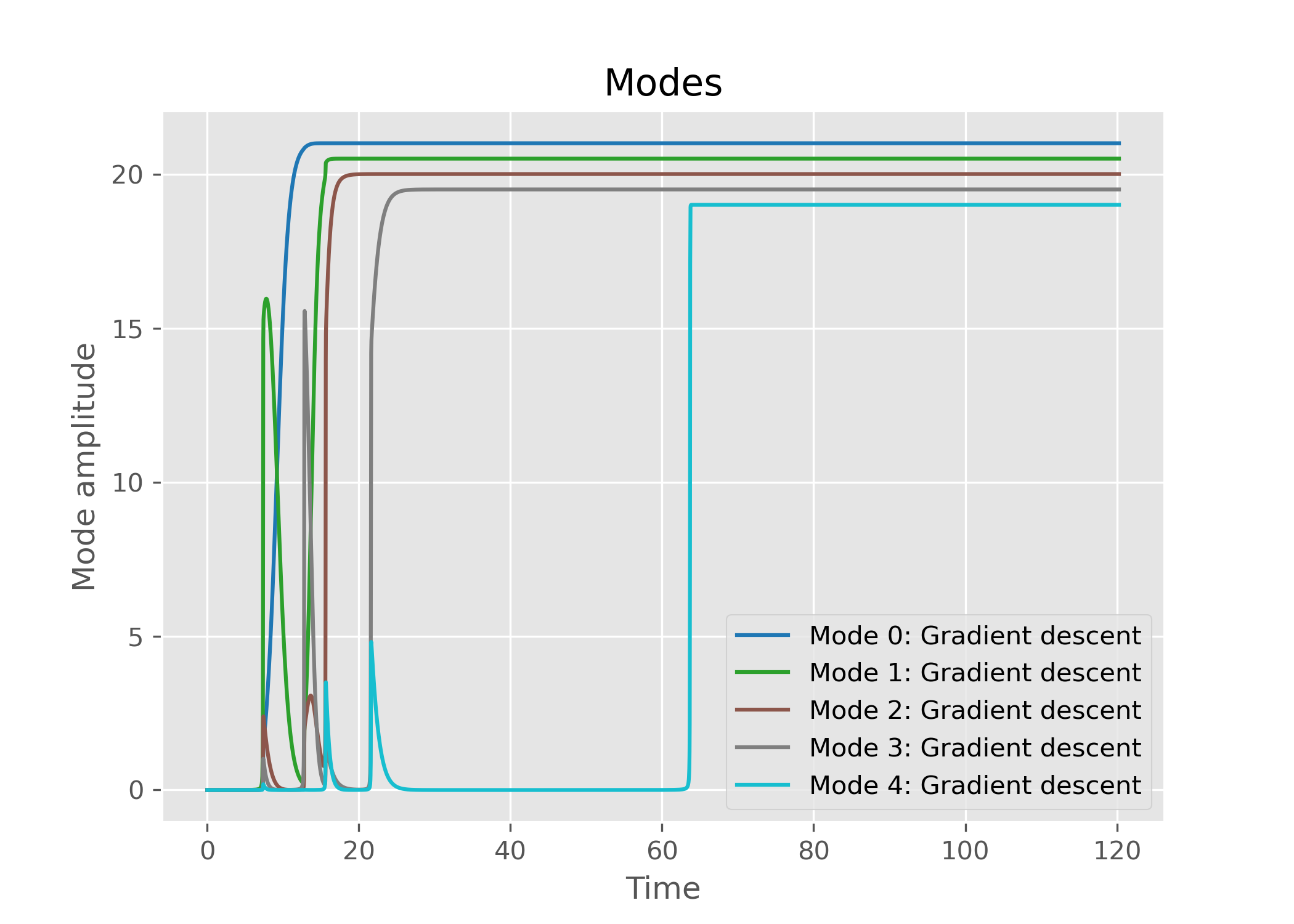}
        \caption{Mode growth with gradient descent}
    \end{subfigure}
    \begin{subfigure}{0.49\textwidth}
        \includegraphics[width=\linewidth]{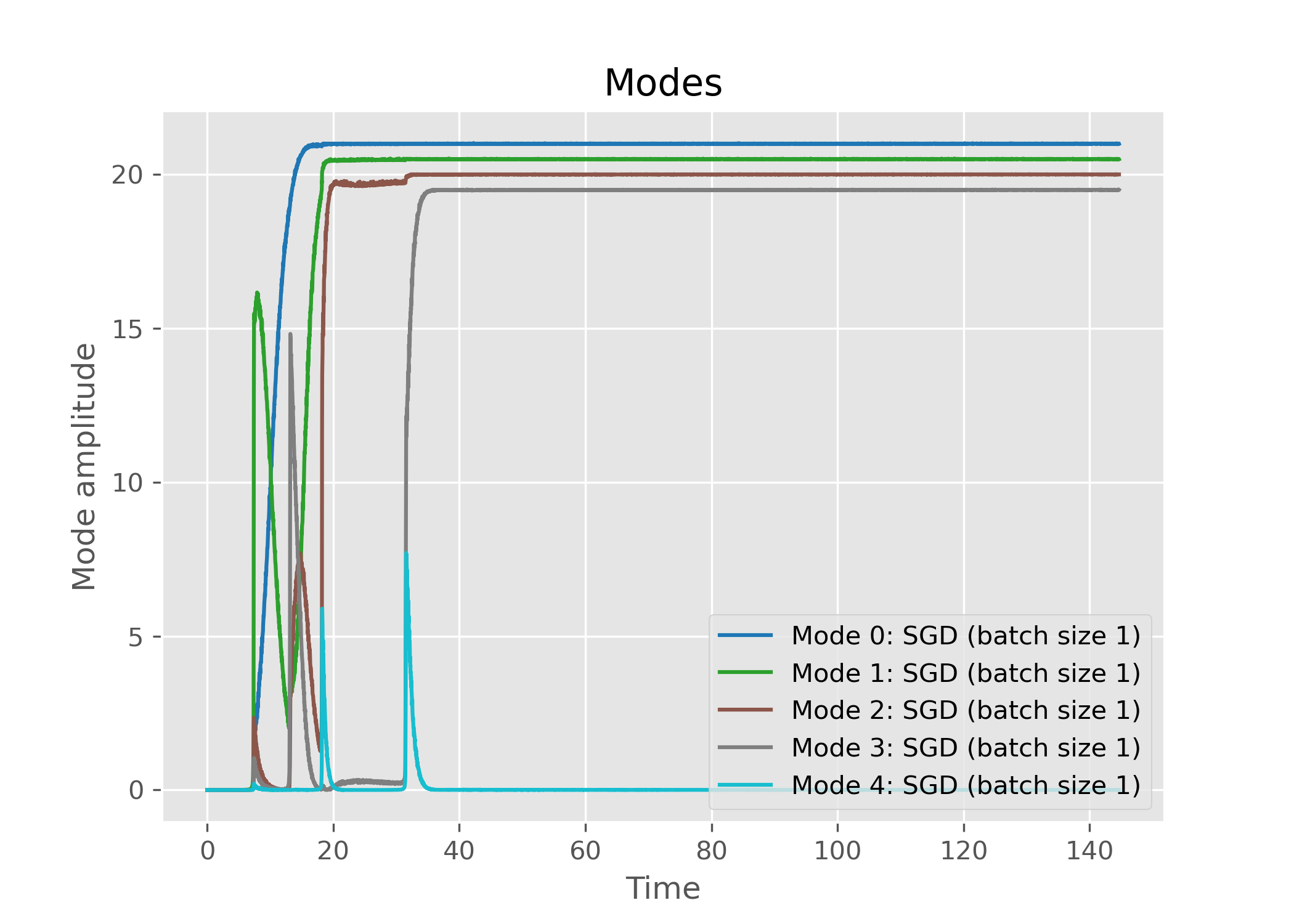}
        \caption{Mode growth with SGD}
    \end{subfigure}
    \caption{\textbf{Saddle-to-saddle dynamics with different optimizers} in a depth-6 linear network, with the sharp changes corresponding to increases in the numerical rank of the network. (a) Train loss plateaus with discrete optimizers (b) Train loss plateaus with numerical simulation of their continuous counterparts (c) Mode growth over training with gradient descent, showing that the 5 singular values of the teacher matrix are learned in descending order of magnitude (d) Mode growth over training for stochastic gradient descent, showing that the stagewise dynamics are retained, but modes take longer to be learned. (See \autoref{app:experimental_setup} for details of the numerical simulations and experiments.)}
    \label{fig:stagewise-intro}
\end{figure}
In this work, we consider a DLN of depth $L$ with small initialization of its weights, i.e., in the feature-learning regime. We take the continuous limit of SGD \ref{subsec:sgd-to-sde} into a stochastic differential equation (SDE) with state-dependent and anisotropic Gaussian noise \ref{eq:sgd-sde}.  We study the dynamics of the mode amplitude during anisotropic Langevin dynamics under the assumption of aligned and balanced weights.

\subsection{Assumptions}
We make the following assumptions:
\begin{itemize}
    \item \textbf{Continuous-time model}: We model SGD by  the SDE in \autoref{eq:sgd-sde}, with state-dependent anisotropic noise.
    \item \textbf{Whitened inputs}: The input distribution $p_X$ is Gaussian with the covariance matrix being the identity. \footnote{This can probably be relaxed by demanding that the input and input-output covariance matrices are diagonalizable in the same basis.}
    \item \textbf{Online learning}: Batches are sampled directly from the data distribution, rather than from a finite dataset. This can be seen as a large-sample limit of the offline, finite-dataset case; \autoref{app:finite-dataset} discusses relaxation to the offline case.
\end{itemize}

\noindent To derive decoupled modewise SDEs, we additionally assume:

\begin{enumerate}[label=\textbf{(A\arabic*)}]
\item\label{assum:balance} \textbf{Balanced weights:} The layer weight matrices satisfy $W_{\ell+1}(t)^\top W_{\ell+1}(t) = W_\ell (t) W_\ell(t)^\top$ for all layers $\ell$ and all times $t$ in training. This condition is preserved under gradient flow.

\item\label{assum:align} \textbf{Aligned modes:} The cross-mode amplitudes vanish, i.e., $w_{\alpha\beta}(t) := u_\alpha^\top W(t) v_\beta = 0$ for all mode indices $\alpha \neq \beta$ and all times $t$ in training. Equivalently, the student $W$ is diagonal in the teacher's singular basis.
\end{enumerate}

\noindent These standard assumptions in the DLN literature, and Appendix~\ref{app:testing-assumptions} tests how well they hold in practice. We refer to \ref{assum:balance}--\ref{assum:align} collectively as the \emph{balance and alignment assumptions}.

\section{Modewise state-dependent SDE for SGD on DLNs}
This section contains results about how the modes evolve under the continuous-time model of \autoref{eq:sgd-sde}. We start by showing that the population loss gradient is a product of the Jacobian and the student-teacher gap $M-W_L\dots W_1$.
\begin{proposition}[Gradient structure]
For each layer \(l\) of a DLN, define the partial products
\[
W_{>l}:=W_L\cdots W_{l+1},\qquad W_{<l}:=W_{l-1}\cdots W_1,
\]
with the convention that an empty product is the identity.
The Jacobian of the end-to-end map $W:=W_L\dots W_1$ with respect to $W_l$ satisfies:
    \begin{equation*}
        J_l: = W_{<l}^{\top} \otimes W_{>l}^{\top}; \quad J_l(x): = x^{\top}W_{<l}^{\top} \otimes W_{>l}^{\top}
    \end{equation*}
The Jacobian with respect to all layers is the block matrix:
\[
\,J(\theta;x)=\big[\,J_1(x)\ \mid\ J_2(x)\ \mid\ \cdots\ \mid\ J_L(x)\,\big]
\]
Let $\Delta:=M -W=M-W_L...W_1$ be the error term between the teacher matrix and the end-to-end linear map $W:=W_L...W_1$ that the network implements. The population gradient of the mean-square error loss function is:
\begin{equation*}
    G_l:= \nabla_{W_l} L =  -W_{>l}^{\top}\Delta W_{<l}^{\top}
\end{equation*}
In vectorized form:
\begin{equation*}
    g_l := \textup{vec}(G_l) =  -(W_{<l} \otimes W_{>l}) \textup{vec}(\Delta) = - J_l^{\top}\delta
\end{equation*}
The gradient vector of the population loss is given by $g := (g_l)_{l}$
\end{proposition}
The gradient is zero if and only if the Jacobian of the implemented linear map $W$ is orthogonal to the teacher-student gap term $\delta$.  It is interesting to observe that the set of critical points associated with a given level set of the loss is highly degenerate. Indeed, the loss and the zero set of the gradient are both invariant under transforming the weight matrices of the hidden layers by some action of the general linear group, i.e., for a hidden layer $l$ of width $d_l$, we have:
$$ W_lP^{-1}_lP_lW_{l-1}=W_lW_{l-1}, \ \text{for} \ P \in GL_{d_l}(\mathbb{R})$$ 
Next, we derive the covariance matrix of the gradient noise. 
\begin{proposition}\label{prop: gn-covariance}
Let \(X\sim\mathcal{N}(0,I_d)\) be the input data, let \(\xi_q\) be the label noise with covariance \(\Sigma_q\) and zero mean, with $\mathbb{E}[\xi_qX^{\top}]=0$, and $\xi_q\perp X$.\footnote{This amounts to modelling aleatoric uncertainty in the output} 
We consider the stacked (vectorized) one-sample gradient noise across all layers and its covariance matrix \(\Sigma(\theta)\in\mathbb{R}^{d\times d}\) (with \(d=\sum_l \dim(\mathrm{vec}\,W^l)\)) such that \(\Sigma(\theta)=\big[\Sigma_{lm}(\theta)\big]_{l,m=1}^L\) is the stacked covariance of the vectorized layerwise gradient noise.\footnote{We use \(\mathrm{vec}(\cdot)\) and the Kronecker product \(\otimes\).} Let $\Delta:=M-W=M-W_L...W_1$ be the gap between the teacher matrix $M$ and the product of the weight matrices $W$.
For all \(l,m\), we have:
\[
\boxed{\;
\Sigma_{lm}(\theta)
\;=\;
\underbrace{J_l^{\top}(I\otimes \Delta)\,(I_{d_0^2}+C)\,(I\otimes\Delta)^\top J_m}_{\text{data-mismatch term}}
\;+\;
\underbrace{(W_{<l} W_{<m}^\top)\;\otimes\;(W_{>l}^\top \Sigma_q W_{>m})}_{\text{label-noise term}}\;,
}
\]
where \(C\) is the commutation matrix satisfying \(C\,\mathrm{vec}(A)=\mathrm{vec}(A^\top)\).
\end{proposition}
A proof of this proposition can be found in the appendix \ref{app:sigma-derivation}. 
The gradient noise covariance matrix is state-dependent and anisotropic. This means that the noise of SGD is structured and depends on the geometry of the loss landscape. In the absence of label noise, we see that the gradient noise covariance is zero at global minima (where $\Delta=0$) and at zeros of the Jacobian of the parameter-function map. 
\smallskip
To see the time-scale separation of feature learning under the SDE model of SGD, we decompose the SDE along the modes of the teacher matrix $M$. 
The next proposition provides a general form for the diffusion of SGD along specific modes. It relies on the same assumptions as the decomposition along modes during gradient flow in \cite{saxe2013exact}.
\begin{proposition}[Mode and cross-mode diffusion]
\label{prop:mode-diffusion-main}
Let $v_{\alpha}$ and $u_{\alpha}$ be a fixed pair of right and left singular vectors (respectively) of the teacher matrix $M$. Define the mode amplitude $w_\alpha(\theta):=u_\alpha^\top W\, v_\alpha$.

The gradient of the mode amplitude with respect to the weight matrix $W_l$ satisfies
\begin{equation*}
    \; \partial_{W^l}w_\alpha
\;=\; \big(W_{>l}^\top u_\alpha\big)\big(W_{<l} v_\alpha\big)^\top.
\quad
\end{equation*}
Based on this, we define the  modewise Jacobian $a_{l,\alpha}$ by
\[
A_{l,\alpha}:=(W_{>l}^\top u_\alpha)\,(W_{<l} v_\alpha)^\top,\qquad a_{l,\alpha}:=\mathrm{vec}\,A_{l,\alpha} =J_l^{\top} (v_{\alpha} \otimes u_{\alpha}),
\qquad a_\alpha:=(a_{1,\alpha};\dots;a_{N,\alpha}).
\]
Under the stacked SDE with batch-size-one noise covariance \(\Sigma(\theta)\), the diffusion of modes and cross-modes amplitude can be written as:
\[
\boxed{\;
D_\alpha(\theta)
= \eta\,a_\alpha(\theta)^\top \Sigma(\theta)\,a_\alpha(\theta),\qquad
D_{\alpha\beta}(\theta)
= \eta\,a_\alpha(\theta)^\top \Sigma(\theta)\,a_\beta(\theta)\;,
}
\]
which determines the one-step mode covariation:
$$\mathbb{E}[\,dw_\alpha\,dw_\beta]= D_{\alpha\beta}(\theta_t)dt.$$
Under the assumptions of whitened inputs, define the Neural Tangent  operator of a DLN at layer $l$ as:
\begin{equation*}
    K_l := J_lJ_l^{\top}
\end{equation*}
In the absence of label noise, using the expression for the gradient noise covariance $\Sigma$, we have a direct relation between the modewise diffusion scalar and the NTK operator of the DLN:
\begin{equation*}
    D_{\alpha\beta} = \eta (v_\alpha^{\top}\otimes u_{\alpha}^{\top})\sum_{l,m} K_l (I\otimes \Delta)\,(I_{d_0^2} + C)\,(I\otimes\Delta)^{\top} K_m (v_{\beta}\otimes u_{\beta}).
\end{equation*}
\end{proposition}
The derivation of the modewise diffusion matrix is in the appendix \ref{app:diffusion-DLN}. We refer to the quantity $\eta a_\alpha(\theta)^\top \Sigma(\theta) a_\alpha(\theta)$ as the empirical or observed diffusion along mode $\alpha$.

The relation between the modewise diffusion scalar and the NTK hints that the diffusion of SGD is sensitive to feature learning. Specifically, we already know that during feature learning the noise of SGD will be sensitive to the directions of the learned feature, given its dependency on $K_l$. However, in the lazy regime during which the NTK is frozen at initialization, the modewise diffusion of SGD will vary only with the teacher-student gap $\Delta$.

The next proposition states a general form for the stochastic dynamics along the modes, which holds in general in the diffusion limit of SGD (i.e., without needing to assume that the weights are balanced and aligned).

\begin{proposition}[General modewise SDE with state dependent noise]
\label{prop:mode-Ito}
Let $\delta:=vec(\Delta)$, let $K_l$ be the NTK operator at layer $l$, let $J$ be the block Jacobian of the DLN, and let $v_{\alpha}$ and $u_{\alpha}$ be a fixed pair of right and left singular vectors (respectively) of the teacher matrix $M$. Given the Itô SDE: 
\[
d\theta_t \;=\; -\,g(\theta_t)\,dt \;+\; \sqrt{\eta}\,\sigma(\theta_t)\,dB_t,
\qquad
\sigma(\theta_t)\sigma(\theta_t)^\top=\Sigma(\theta_t).
\]
the scalar modewise amplitude process $w_\alpha(t)$ obeys the SDE
\[
\boxed{\;
dw_\alpha(t)
=\mu_\alpha(\theta_t)\,dt
+\sqrt{\eta}\,(v_{\alpha}^{\top}\otimes u_{\alpha}^{{\top}})J \sigma(\theta_t)\,dB_t,
\qquad
\mu_\alpha(\theta)
=-(v_{\alpha}^{\top}\otimes u_{\alpha}^{\top})\sum_{l=1}^L\!K_l \delta
+\frac{\eta}{2}\,\mathrm{tr}\!\left(\Sigma(\theta)\,\nabla^2 w_\alpha(\theta)\right),
\;}
\]
where for $l>m$ (with $l,m$ swapped for $l<m$):  $$\nabla^2_{l,m} w_\alpha(\theta_t) = (W_{l-1:m+1} \otimes W_{>l}^{\top}u_\alpha)(v_{\alpha}^{\top}W_{<m}^{\top}\otimes I); \quad W_{l-1:m+1} := W_{l-1}...W_{m+1}$$ is the Hessian of the modes amplitudes in the stacked coordinates of $W_l$ and $W_m$, whose
diagonal blocks ($l=m$) vanish.
\end{proposition}
The drift term $\mu_{\alpha}$ is a combination of two drifts. The first term, which is a sum on the NTK and the teacher-student gap $K_l\delta$, is the usual gradient-induced drift term which we also find in gradient flow. This term governs the evolution of the feature directions. It vanishes when the teacher–student gap $\delta$ has no projection in directions orthogonal to the current subspace of features. In that case, adjusting the feature directions cannot further reduce the gap, so only the singular values evolve. The second drift term is a drift induced by the noise of SGD, which is a consequence of taking the Itô derivative of a mode amplitude.  Interestingly, this drift induced by noise is a scalar product between the Hessian of the mode amplitude and the gradient noise covariance matrix. In particular, this drift induced by noise will be zero when the noise is orthogonal to the flat directions of the mode amplitude, i.e., when no learning of the mode happens.

The derivation in proposition \ref{prop:mode-Ito} is in Appendix \ref{app:sde-ald-dlns}. In \autoref{fig:stagewise-intro}, we report the modewise dynamics of SGD and its continuous limit. Similarly to gradient flow, modes are learned in a decreasing order of magnitude in the feature learning regime. The main difference is that the time-scale of learning is not the same for SGD and its continuous limit to state-dependent noise, as it is typically slower than gradient descent. (Further details of the setup used for experiments are given in \autoref{app:experimental_setup}.)
\\
\\
\textit{Remark} If we replace the state-dependent SDE of SGD with a Langevin SDE with isotropic and homogeneous noise, the modewise diffusion terms become:
\begin{align*}
D_\alpha(\theta) & \;=\;\eta\,\sigma^2\,\|a_\alpha(\theta)\|^2 = \eta\sigma^2 (v_{\alpha}^{\top} \otimes u_{\alpha}^{\top})\sum_l K_l(v_{\alpha} \otimes u_{\alpha}),\\
D_{\alpha\beta}(\theta)& \;=\;\eta\,\sigma^2\,\langle a_\alpha(\theta),\,a_\beta(\theta)\rangle = \eta\sigma^2 (v_{\alpha}^{\top} \otimes u_{\alpha}^{\top})\sum_l K_l (v_{\beta} \otimes u_{\beta})
\end{align*}
and the Itô-induced drift is simply proportional to the trace of the Hessian of the mode amplitude. Interestingly, even though the noise of Langevin dynamics is state-independent and isotropic, the modewise diffusion is in general state-dependent as it depends on the NTK $K_l$. If the NTK is frozen at initialization, then the modewise diffusion of Langevin will be isotropic and state-independent; however, during the feature learning regime, the modewise diffusion of Langevin will be structured by the NTK and the directions and amplitude of the features being learned.

We study the solutions of the modewise SDE of proposition \ref{prop:mode-Ito} analytically by assuming that the mode are aligned and that the weights are balanced, conditions that are often assumed and satisfied when studying DLN training dynamics. 
\begin{proposition}[Modewise decoupled SDEs for aligned modes and balanced weights]\label{prop:modewise-SDE-balanced}
Assume the conditions of the previous proposition, and additionally the assumptions \ref{assum:balance} and \ref{assum:align} (balance and alignment).
Let $\beta:=\frac{\eta}{b}$ be the ratio of the learning rate and the batch size. Then, each mode amplitude $w_\alpha$ evolves independently according to the scalar It\^o SDE
\begin{equation}\label{eq:modewise-SDE}
\boxed{\quad
dw_\alpha
= \Big[\,\mu_\alpha^{\rm grad}(w_\alpha)\;+\;\mu_\alpha^{\rm Ito}(w_\alpha)\,\Big]\,dt
\;+\;\sqrt{D_\alpha(w_\alpha)}\,dB_{\alpha,t},
\quad}
\end{equation}
with drift components
\begin{align}
\mu_\alpha^{\rm grad}(w_\alpha)
&= (s_\alpha - w_\alpha)\,L\,w_\alpha^{\frac{2(L-1)}{L}},
\label{eq:grad-drift}\\[2mm]
\mu_\alpha^{\rm Ito}(w_\alpha)
&= \beta\,(s_\alpha - w_\alpha)\,L(L-1)\,w_\alpha^{\frac{2(L-2)}{L}}
\;+\;\beta\,L(L-1)\,w_\alpha^{\frac{2(L-2)}{L}}\;\sigma_q,
\label{eq:ito-drift}
\end{align}
and diffusion (mode-diagonal) coefficient
\begin{equation}\label{eq:diffusion}
D_\alpha(w_\alpha)
=\beta\,L^{2}\Big(2\,(s_\alpha-w_\alpha)^{2}+\sigma_q^2\Big)\,w_\alpha^{\frac{4(L-1)}{L}}.
\end{equation}
In the above, $B_{\alpha,t}$ is a Wiener process.
\end{proposition}
The derivation of the proposition is in \autoref{app:proof-of-modewise-SDE-balanced}.

\autoref{fig:summary} shows that the formula in \autoref{eq:diffusion} is a good match for the expression for $D_\alpha$ in Proposition~\ref{prop:mode-diffusion-main}. \autoref{app:testing-assumptions}, in which we run experiments with various hyperparameters, establishes the diffusion scales linearly with learning rate and inversely with batch size, as \autoref{eq:diffusion} predicts.

In the absence of label noise, we see that the diffusion along a particular mode vanishes once the mode is fully learned (i.e., once $w_\alpha=s_\alpha$). This observation is compatible with degeneracies in the loss landscape causing SGD noise to continuously tend to zero as it approaches a degenerate critical locus of the loss \cite{corlouerSGDsticky}. In the presence of label noise, we recover Langevin dynamics with anisotropic and state-independent noise once a mode has been learned.

It is interesting to note that SGD noise does not seem to fundamentally alter the training dynamics of the DLN that occur with gradient flow. Indeed, the structured noise experimentally appears to only change the speed at which modes are learned, and we still observe a separation of time scales for the learning of the modes (see \autoref{fig:stagewise-intro}). SGD is going through a saddle-to-saddle dynamics in which the time spent between saddles depends on the size of each singular value via the drifts and diffusion induced by gradient and SGD noise.

\section{State dependent noise predicts feature learning}
A feature is fully learned once the mode amplitude $w_{\alpha}$ reaches the corresponding singular value $s_{\alpha}$ of the teacher matrix $M$. An interesting corollary of the system of one-dimensional SDEs in Proposition~\ref{prop:modewise-SDE-balanced} is that the state-dependent noise of SGD can predict when a feature will be fully learned by tracking peaks of the corresponding modewise diffusion coefficient.

We consider the modewise diffusion coefficient from Proposition~\ref{prop:modewise-SDE-balanced}:
\begin{equation}\label{eq:diffusion2}
D_\alpha(w_\alpha)
=\beta\,L^{2}\Big(2\,(s_\alpha-w_\alpha)^{2}+\sigma_q^2\Big)\,
w_\alpha^{\frac{4(L-1)}{L}}.
\end{equation}
Maximizing~\eqref{eq:diffusion2} with respect to $w_\alpha>0$ gives the stationary condition
\[
\frac{dD_\alpha}{dw_\alpha}=0
\;\;\Longleftrightarrow\;\;
(2a+4)(w_\alpha-s_\alpha)^{2}
+4s_\alpha(w_\alpha-s_\alpha)
+a\,\sigma_q^{2}=0,
\qquad a=\frac{4(L-1)}{L}.
\]
Solving for $w_\alpha$ yields the critical point
\begin{equation}\label{eq:wstar}
w_\alpha^\star
= 
\frac{(a+1)\,s_\alpha
-\sqrt{\,s_\alpha^{2}-\frac{a(a+2)}{2}\,\sigma_q^{2}\,}}
{\,a+2\,}.
\end{equation}
This maximum exists if and only if the discriminant is non-negative, that is
\begin{equation}\label{eq:sigmaq_condition}
\sigma_q^{2}
\;\le\;
\frac{2s_\alpha^{2}}{\,a(a+2)\,}
\;=\;
\frac{s_\alpha^{2}L^{2}}{\,4(L-1)(3L-2)\,}.
\end{equation}
Whenever~\eqref{eq:sigmaq_condition} holds, the maximizer satisfies 
$w_\alpha^\star < s_\alpha$, meaning that the diffusion coefficient 
peaks before the mode is fully learned. 

\subsection{Experimental results}
\begin{figure}
 \begin{subfigure}{0.49\textwidth}
        \includegraphics[ height=8cm, keepaspectratio]{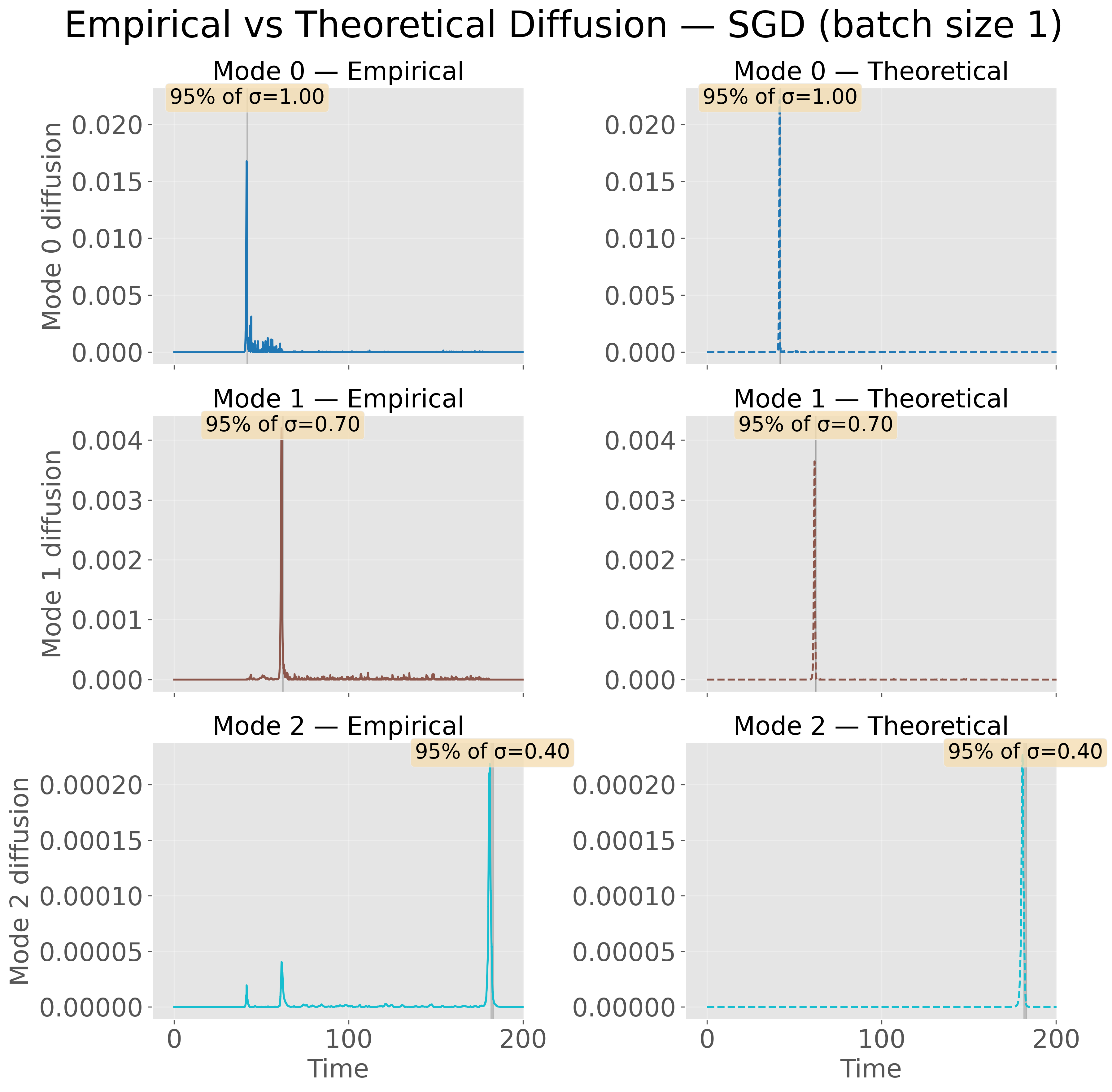}
        \caption{SGD}
    \end{subfigure}
    \begin{subfigure}{0.49\textwidth}
        \includegraphics[height=8cm, keepaspectratio]{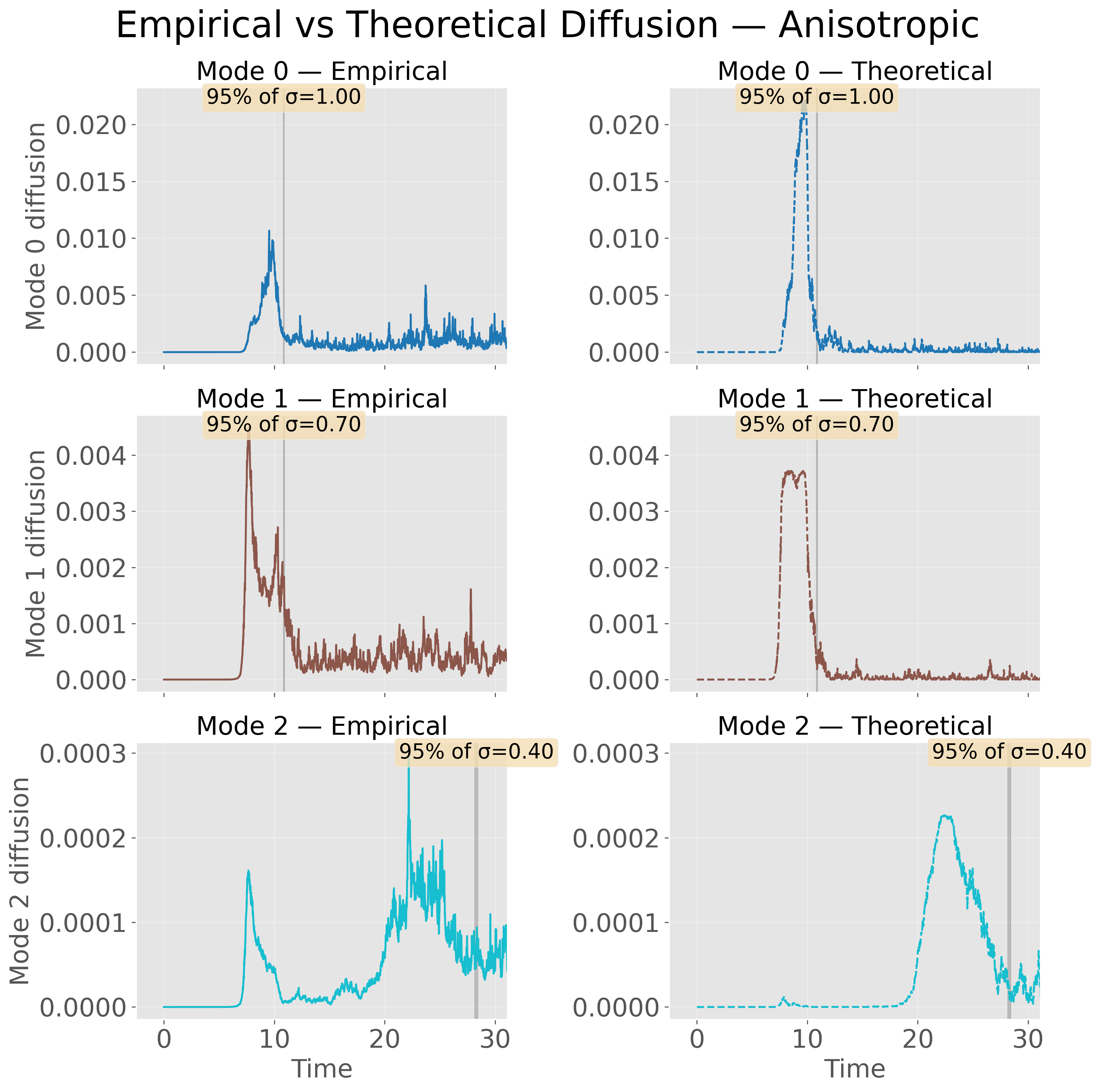}
        \caption{Simulation of stochasic gradient flow, with anisotropic noise}
    \end{subfigure}
    \caption{\textbf{Comparison of diffusion along modes with mode amplitude}, for (a) SGD and (b) an Euler-Maruyama simulation of  anisotropic Langevin dynamics. In each column, the empirical value of $\eta a_\alpha(\theta)^\top \Sigma(\theta) a_\alpha(\theta)$ is shown on the left and Proposition~\ref{prop:modewise-SDE-balanced}'s theoretical prediction for $D_\alpha(\theta)$ is shown on the right. The shaded bands show the time that the corresponding mode is learned. In agreement with the theoretical prediction, both have peaks in the diffusion before the mode is learned.}
    \label{fig:noise-predict-feature}
\end{figure}

In \autoref{fig:noise-predict-feature}, we observe  that the variance of state-dependent noise of SGD peaks before each mode is learned, and that the theoretical predictions, agreeing with the previous section. Additionally, the empirical diffusion coefficient for each mode, using the quantity $D_\alpha(\theta)=\eta a_\alpha(\theta)^\top \Sigma(\theta) a_\alpha(\theta)$ from Proposition~\ref{prop:mode-diffusion-main}, has similar behavior to the analytic form given in Proposition~\ref{prop:modewise-SDE-balanced} for the aligned and balanced case. Note that this similarity exists even though balance and alignment are not fully satisfied (see \autoref{app:testing-assumptions}).

\autoref{fig:summary} also shows how the critical points in \autoref{eq:wstar} correspond to the times when modes are learned. 

\autoref{fig:noise-predict-feature} again shows that the discrete online SGD has a slower timescale of learning than the simulation of anisotropic Langevin dynamics, taking 200 units of time to learn all modes as opposed to 30. It also has more abrupt peaks in the modewise diffusion.

\section{Marginal modewise stationary distribution for anisotropic Langevin dynamics}
We study the end of training distribution by approximating the stationary distribution of the mode amplitudes using the decoupled modewise SDEs and their induced Fokker-Planck equations.
\begin{proposition}[Fokker--Planck equations]
\label{prop:FP}
Let $\theta_t$ be a solution of the SDE \ref{eq:sgd-sde}, and let $p(\theta,t)$ be the density of $\theta_t$. Then $p(\theta,t)$ satisfies the following Fokker-Planck equation:
\[
\;
\partial_t p(\theta,t)
=\nabla_\theta\!\cdot\!\Big(g(\theta)\,p(\theta,t)\Big)
+\frac{\eta}{2}\sum_{i,j}\partial_{\theta_i}\partial_{\theta_j}\Big(\Sigma_{ij}(\theta)\,p(\theta,t)\Big).
\;
\]
Given no cross-modes amplitude and balanced weights, the mode amplitude $w_\alpha$ satisfies the Itô SDE:
$dw_\alpha=\mu_\alpha(w_\alpha)\,dt+\sqrt{D_\alpha(w_\alpha)}\,d\beta_t$, and thus the modewise probability density $p_\alpha(w_\alpha,t)$ satisfies the following 1D Fokker-Planck equation:
\[
\boxed{\;
\partial_t p_\alpha(w_\alpha,t)
= -\partial_w\!\big(\mu_\alpha(w_\alpha)\,p_\alpha(w_\alpha,t)\big)
+\tfrac{1}{2}\,\partial_w^2\!\big(D_\alpha(w_\alpha)\,p_\alpha(w_\alpha,t)\big).
\;}
\]
\end{proposition}

\begin{proposition}[Modewise stationary law under detailed balance]\label{prop:modewise-stationary}
Consider the scalar It\^o SDE under the assumptions of no cross-mode amplitude and balanced weights. The mode amplitude $w_\alpha$ satisfies:
\begin{equation}\label{eq:mode-sde}
dw_\alpha \;=\; \mu_\alpha(w_\alpha)\,dt \;+\; \sqrt{D_\alpha(w_\alpha)}\,dB_{\alpha,t},
\end{equation}
with drift and diffusion given (from Proposition~6.4) by
\begin{align}
\mu_\alpha(w)
&= \mu_\alpha^{\rm grad}(w) + \mu_\alpha^{\rm Ito}(w),\\
\mu_\alpha^{\rm grad}(w)
&= L\,w^{\frac{2(L-1)}{L}}\,(s_\alpha - w),\\
\mu_\alpha^{\rm Ito}(w)
&= \beta\,L(L-1)\,w^{\frac{2(L-2)}{L}}\!\left[(s_\alpha - w) + \tfrac12\,\sigma_q\right],\\
D_\alpha(w)
&= \beta\,L^2\!\left(2(s_\alpha - w)^2 + \sigma_q^2\right) w^{\frac{4(L-1)}{L}}.
\end{align}
Here $L\in\mathbb{N}$ is the depth, $s_\alpha>0$ is the teacher singular value for mode $\alpha$, $\eta>0$ is the learning rate, $\beta>0$ is the temperature, and $\sigma_q\ge 0$ is the label noise scale. Assume  \textit{detailed balance} for the Fokker--Planck equation induced by \eqref{eq:mode-sde} and $D_{\alpha}(w_{\alpha}) > 0$. Then any stationary density $p_\alpha^\star$ satisfies
\begin{equation}\label{eq:general-stationary}
p_\alpha^\star(w)
\;\propto\;
\frac{1}{D_\alpha(w)}\,
\exp\!\left(\int_{0}^w \frac{2\,\mu_\alpha(z)}{D_\alpha(z)}\,dz\right).
\end{equation}
Moreover, we have the following:
\begin{enumerate}
\item[(i)] \textbf{No label noise} ($\sigma_q=0$): $D_\alpha(w)\sim Q\,(s_\alpha-w)^2$ and $\mu_\alpha(w)\sim q\,(s_\alpha-w)$ as $w\uparrow s_\alpha$ with constants $Q,q>0$. The density \eqref{eq:general-stationary} is non-normalizable unless it collapses to a Dirac mass, hence
\[
p_\alpha^\star \;=\; \delta(w-s_\alpha).
\]
\item[(ii)] \textbf{With label noise} ($\sigma_q>0$): there is a unique smooth stationary density peaked near the zero of $\mu_\alpha(w)$. One has the small-$\eta$ expansions
\begin{align}
w_\alpha^\star - s_\alpha
&= \frac{\eta}{2B}\,\,(L-1)\,s_\alpha^{-2/L}\,\sigma_q \;+\; O(\eta^2),\\
\operatorname{Var}_{p_\alpha^\star}(w_\alpha)
&= -\frac{D_\alpha(s_\alpha)}{2\,\mu_\alpha'(s_\alpha)} \;+\; O(\eta^2)
= \frac{\eta\,L}{2B}\,\sigma_q^2\,s_\alpha^{\frac{2(L-1)}{L}} \;+\; O(\eta^2),
\end{align}
so $p_\alpha^\star$ is approximately Gaussian with mean $s_\alpha + O(\eta)$ (slightly \emph{above} $s_\alpha$ due to the It\^o drift) and $O(\eta)$ variance.
\end{enumerate}
\end{proposition}
The proof of this proposition can be found in the appendix \ref{sec:proof stationary-law}. 
\\
\\
The latter proposition shows that in the absence of label noise, the modewise distribution of parameters is similar for SGD and GD, i.e., a Dirac at a specific mode. In the presence of label noise, SGD appears to approximate a Boltzmann distribution at finite time, although we suspect that it converges to a Dirac in the long run.

\subsection{Experimental results}

\begin{figure}
    \centering
    \begin{subfigure}[t]{0.45\textwidth}
        \includegraphics[height=6cm]{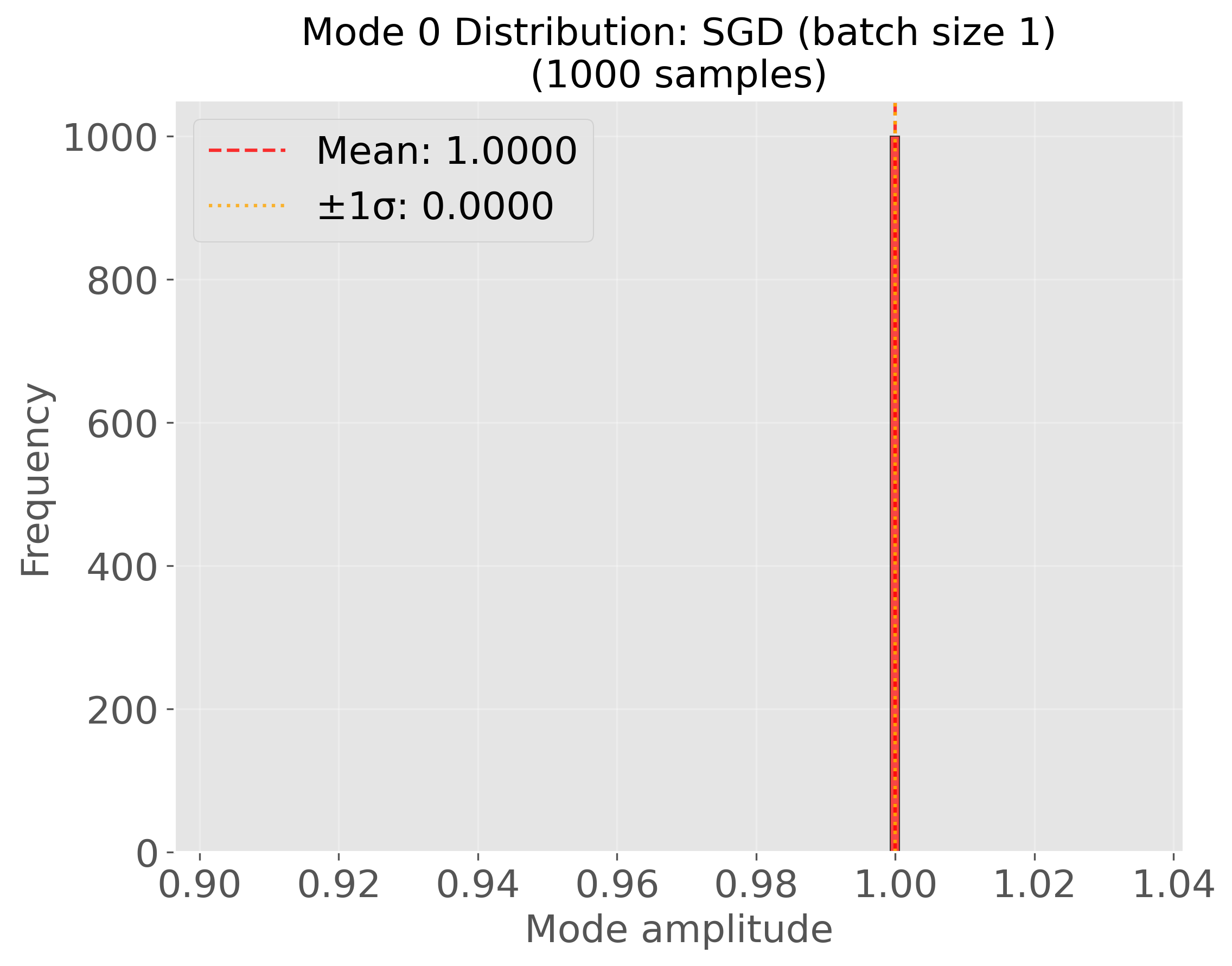}
        \caption{Online SGD}
        \label{fig:distribution-convergence-sgd}
    \end{subfigure}
     \begin{subfigure}[t]{0.45\textwidth}
        \includegraphics[height=6cm]{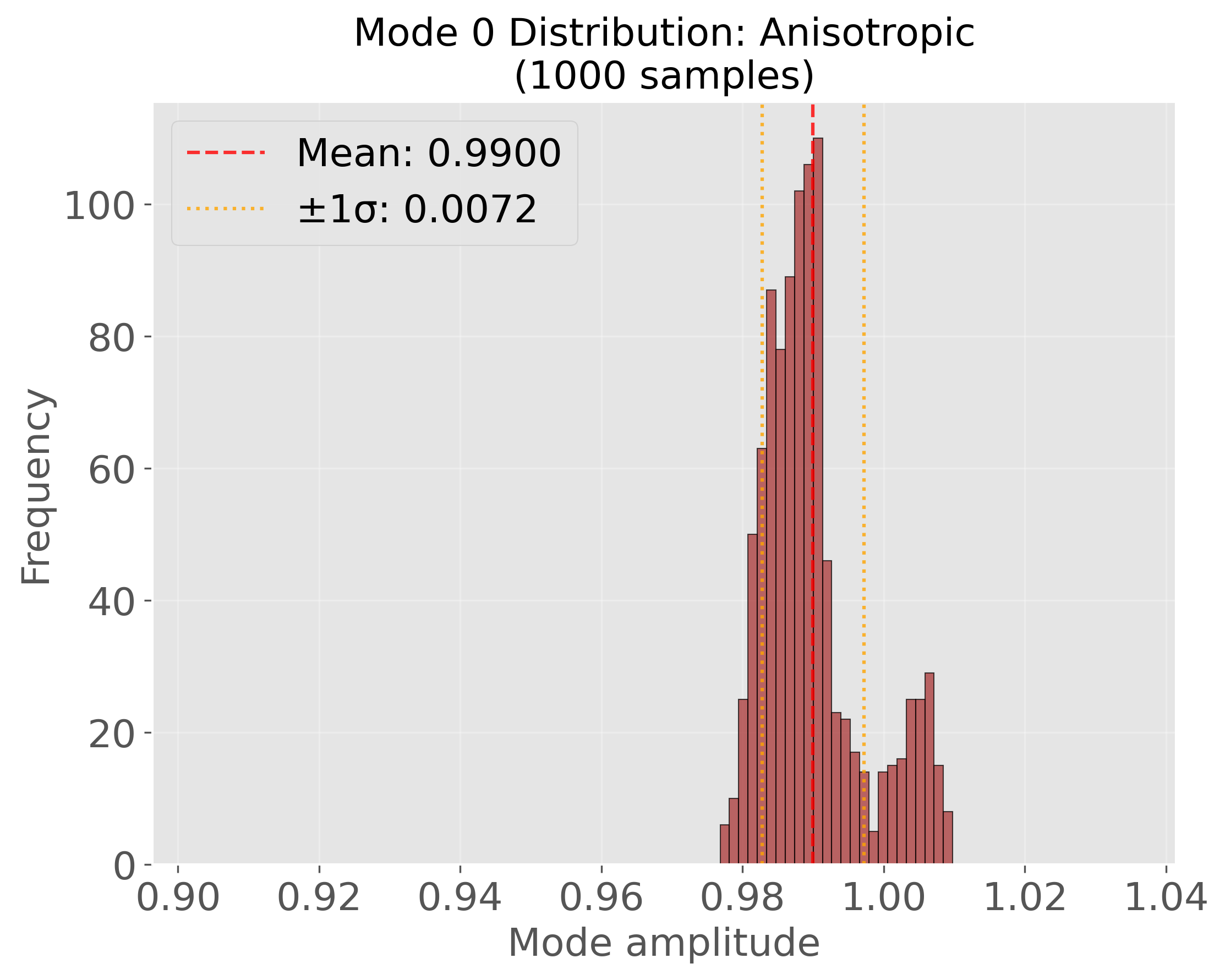}
        \caption{Euler-Maruyama simulation of  \autoref{eq:sgd-sde}, with anisotropic and state-dependent noise}
        \label{fig:distribution-convergence-aniso}
    \end{subfigure}
    \begin{subfigure}[t]{0.45\textwidth}
        \includegraphics[height=6cm]{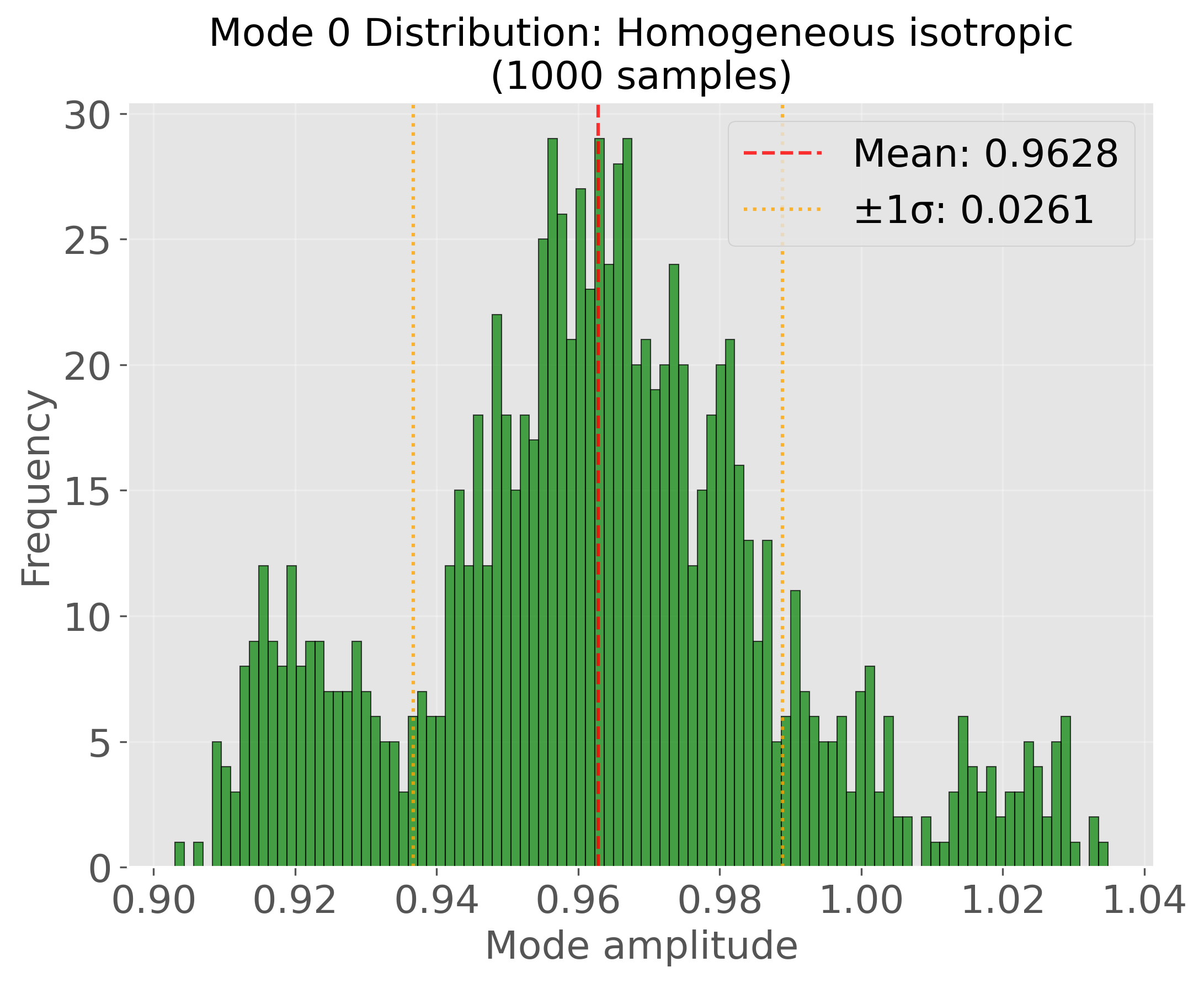}
        \caption{Isotropic Gaussian noise}
    \end{subfigure}
    
    \caption{\textbf{Comparison of the empirical distributions} of the amplitude of the first mode $w_0$ at the end of training for SGD, anisotropic Gaussian, and isotropic Gaussian noise. In the absence of label noise, SGD (a) concentrates entirely on the value of the top singular value of the teacher matrix, but anisotropic noise (b) does not have this behavior. However, the variance of the distribution for anisotropic noise is lower than for isotropic noise (c), and is thus more similar to SGD.}
    \label{fig:distribution-convergence}
\end{figure}

In \autoref{fig:distribution-convergence}, we use histograms to visualize the distribution of the amplitude of the largest mode at the end of training.  SGD, shown in \autoref{fig:distribution-convergence-sgd} sharply peaks around the teacher's mode amplitude (1.0). This matches the theoretical prediction of Proposition~\ref{prop:modewise-stationary}.

However, the Euler-Maruayama discretization of the anisotropic Langevin dynamics SDE (\autoref{eq:sgd-sde}) exhibits a less sharply peaked distribution (\autoref{fig:distribution-convergence-aniso}) than SGD does. We conjecture that the gap between the observed end-of-training distribution and the prediction of a Dirac distribution is because of discretization error; \autoref{app:disc-error} provides evidence for this by showing that the variance of the mode amplitude's distribution reduces when the simulation uses finer time steps.

\begin{figure}
    \centering
    \includegraphics[width=0.5\linewidth]{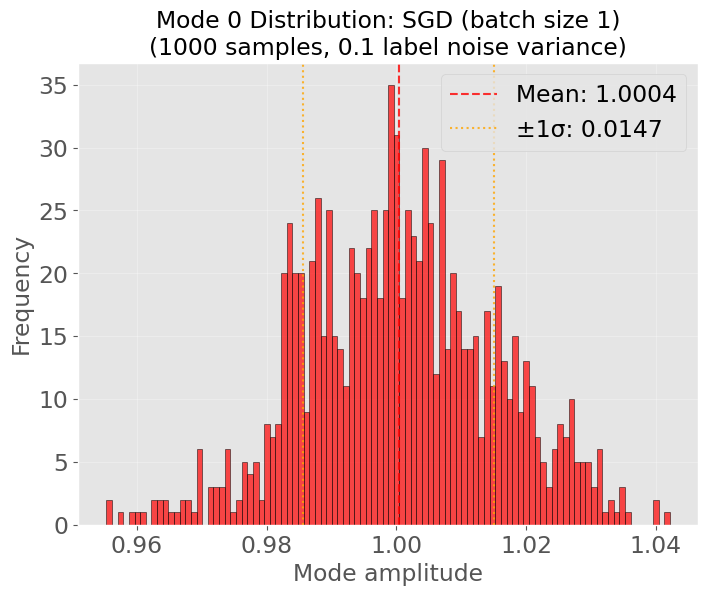}
    \caption{\textbf{End-of-training distribution in the presence of label noise (variance 0.1)} of the amplitude of the first mode $w_0$ for SGD. The distribution changes from being concentrated at a point to having greater variance.}
    \label{fig:distribution-convergence-label-noise}
\end{figure}
\autoref{fig:distribution-convergence-label-noise} shows the same histogram in the presence of label noise for SGD. In agreement with the prediction of part (ii) of Proposition~\ref{prop:modewise-stationary}, the distribution is approximately Gaussian and is no longer concentrated at a point.

\section{Discussion}
We derive the training dynamics of stochastic gradient descent in deep linear networks with balanced weights, aligned modes of the teacher matrix, and whitened inputs. We extended previous analyses of gradient flow in DLNs \citep{saxe2013exact} to stochastic Langevin dynamics as a model of SGD. Furthermore, we considered an anisotropic and state-independent gradient noise covariance matrix of SGD. We gave an analytic expression of the gradient noise covariance matrix in deep linear networks and its behavior along specific modes of the DLN. We found that the modewise diffusion of SGD precedes the time at which a feature is going to be fully learned during the growth of mode amplitude away from its initialization. This observation shows that stochasticity encodes information about the progression of feature learning. Finally, we showed that the stationary modewise distribution of the stochastic Langevin process approaches that of discrete-time SGD (which is the same as GD in the absence of label noise), concentrating around the teacher singular values.

\paragraph{Limitations} First of all, we have modeled SGD with a continuous limit, ignoring the effect of a finite learning rate. However, this assumption might be overcome by using an effective potential instead of the loss function, using a \textit{central flow} term accounting for the oscillations of gradient descent induced by a finite learning rate \cite{cohen2024understanding}. We also assumed that the data is i.i.d. and that SGD noise is not heavy-tailed. Further work could look into heavy-tailedness by using specific data distribution on which heavy-tailedness is important \cite{gurbuzbalaban2021heavy}. We also assumed that the weights are balanced and aligned assumptions, which are common when studying DLNs but the assumption of modes alignment is not always accurate (see figure \ref{fig:cross_modes}). These assumptions are important, and further work could relax them to extend our analyses to take into account cross-modes interactions.
\paragraph{Future directions.}  
The framework developed here opens several avenues for further investigation. Beyond the linear setting, we would like to study how mode-wise diffusion manifests in non-linear architectures, such as two-layer ReLU networks, and to what extent we can track stagewise feature learning. Another direction would be to investigate the Golden Path hypothesis. This hypothesis states that in the online learning regime, in which new, fresh batches are sampled at each time step, stochasticity does not affect generalization and the function selected by the training process and instead is mere computational convenience \cite{vyas2023beyond}. While we have seen that SGD noise carries information about when a new feature is going to be learned, it might in itself not matter for learning a particular feature, as gradient flow is able to learn the same features in the same order as anisotropic Langevin dynamics, and the loss curves have similar qualitative stagewise behavior. Clarifying under which conditions the golden path hypothesis holds would be important, as it would enable us to simplify the theoretical analysis of deep neural networks' training dynamics by restricting it to the study of gradient flow.
\section{Acknowledgments} We used ChatGPT and Claude as tools to assist in mathematical derivations and checking intermediate calculations. All proofs were verified by the first author. ChatGPT and Claude were used to assist with writing, editing, and LaTeX formatting. This work was funded by the Pivotal fellowship. 
\bibliography{bibliography}
\appendix
\section{Setup}
\label{app:finite-dataset}
\subsection{Gradient noise}
Let $B_k$ be a batch of $b$ independent samples from the data distribution $p_{\mathcal{X}\times \mathcal{Y}}$. Define the gradient batch noise as:
\begin{align*}
    \xi(B_k) & := g(\theta_k) - g_{B_k}(\theta_k)\\
\end{align*}
The batch gradient noise covariance matrix is defined as:
\begin{align*}
    \Sigma_{b} & := \mathbb{E}_{B_k}[\xi(B_k)\xi(B_k)^\top]
\end{align*}
Where the expectation is taken over all possible batches of size $b$.
Let $g_i$ be the 1-sample gradient of the loss function $\ell(\theta; y_i, x_i)$ over $N$ samples $(x_i, y_i)$. The empirical gradient noise covariance matrix is defined as:
\begin{align*}
    \hat{\Sigma}_N & := \frac{1}{N}\sum_{i=1}^N g_i g_i^\top - g_N g_N^\top
\end{align*}

Note that, using independence of batch sampling, the covariance matrix for batches of size $b>1$ is a scalar multiple of the covariance matrix in the 1-sample ($b=1$) case:
\begin{equation*}
    \Sigma_{b} = \operatorname{Cov}\!\left(\frac{1}{b}\sum_{i=1}^b g_i\right) = \frac{1}{b^2}\sum_{i =1}^b\operatorname{Cov}(g_i) = \frac{1}{b}\Sigma_1
\end{equation*}

Since batch gradient and 1-sample gradient, as well as their noise covariance matrices, are closely related, we will only consider the 1-sample gradient and noise covariance matrix (denoted $\Sigma$). Let $g(\theta_k; Y_k, X_k)$ be the 1-sample gradient at the iterate $k$. The SGD update rule can be written as a drift term and a noise term:
\begin{align*}
    \Delta\theta_{k+1} &: = \theta_{k+1} - \theta_k = - \eta_k g_N(\theta_k) + \eta_k \xi_k \\
    \xi_k & := g_N(\theta_k) -g(\theta_k;  Y_k, X_k) 
\end{align*}

\subsection{Continuous limit of SGD with constant learning rate}
\label{app:sgd-to-sde}
Let $\theta_k$ be the parameter at iterate $k$ satisfying the SGD update rule. Let $B_t$ be a Wiener process. We want to understand the conditions under which the parameter $\theta_k$ is the solution of the Euler-Maruyama discretization of the following SDE:
\begin{align*}
\label{eqn:ALD}
d\theta(t) = - g_N(\theta, t) dt + \sqrt{\eta\Sigma(\theta, t)} dB(t)
\end{align*}
I.e., we want: $\theta(\eta k) = \theta_k$ for all $k$ and $\theta(t)$ is a solution of the latter SDE.
Iterating the SGD update rule, we have:
\begin{align*}
\theta_k = \theta_0 - \sum_{i=0}^{k-1} \eta g_N(\theta_i) + \sum_{i=0}^{k-1} \eta \xi_i
\end{align*}
Let $t = \eta k$ and $\Delta t = \eta$. We have the following:
\begin{align*}
\theta_{\frac{t}{\eta}} = \theta_0 - \sum_{i=0}^{\frac{t}{\eta}-1} \Delta t g_N(\theta_i) + \sqrt{\eta}\sum_{i=0}^{\frac{t}{\eta}-1} \sqrt{\Delta t}\xi(\theta_i)
\end{align*}
For a given $t$ we have:
\begin{align*}
M(t) & := \sqrt{\eta}\sum_{i=0}^{\frac{t}{\eta}-1} \xi(\theta, t_i)\ \ \implies \
\text{Cov}(M(t_i))  =   \sqrt{\eta}\frac{t}{\eta}\Sigma = \sqrt{\eta}t\Sigma \quad \text{if all $\xi$ have same covariance matrix}
\end{align*}
Let $\mathcal{F}_k$ be a filtration adapted to $\xi_k$. Since $\mathbb{E}[\xi_k|\mathcal{F}_k] = 0$, the process $M(t)$ is a martingale. If there exists a continuous, symmetric, positive semi-definite matrix $\Sigma(s)$ such that, uniformly:
\begin{align*}
\eta \sum_{k=0}^{\frac{t}{\eta}-1} \mathbb{E}[\xi_k\xi_k^{\intercal}|\mathcal{F}k] \to_p \int_0^t \Sigma(s) ds
\end{align*}
Assume that the noise $\xi_k$ satisfies the Lindeberg condition:
\begin{align*}
\forall \epsilon > 0, \quad \lim_{\eta\to 0} \eta\sum_{k=0}^{\frac{t}{\eta}-1} \mathbb{E}[||\xi_k||^21_{||\xi_k|| > \epsilon/\sqrt{\eta}}|\mathcal{F}_k] = 0
\end{align*}
Under these conditions, when taking $\eta \to 0$, the functional central limit theorem ensures that the martingale $M(t)$ converges to a Wiener process with covariance $\Sigma(\theta, t)$ and the process $\theta(t)$ whose Euler-Maruyama discretization is $\theta_k$ satisfies the SDE \ref{eq:sgd-sde}. To model SGD with the SDE \ref{eq:sgd-sde}, one must verify that all the conditions for applying the FCLT are satisfied and take the limit $\eta\to 0$.
\section{Derivation of the Gradient--Noise Covariance \texorpdfstring{$\Sigma_{lm}$}{Σlm}}
\label{app:sigma-derivation}

In this section, we derive the covariance of the per-sample gradient noise in a deep linear network (DLN). This corresponds to proposition \ref{prop: gn-covariance} in the main text.

\subsection*{Assumptions and Setup}

We consider a depth-$L$ deep linear network with weight matrices
\[
W_i\in\mathbb{R}^{d_i}, \ i\in \{1,...L\}
\]
and end-to-end map
\[
W := W_L W_{L-1}\cdots W_1 \in\mathbb{R}^{d_L\times d_0}
\]
The data $(Y,X)\in\mathbb{R}^{d_L\times d_0}$ is generated from a teacher model with additive label noise:
\[
Y = M X + \xi_q,
\]
where
\begin{itemize}
    \item The inputs $X \in \mathbb{R}^{d_0}$ are i.i.d.\ whitened Gaussian, $X \sim \mathcal N(0,I_{d_0})$.
    \item The teacher map is $M \in \mathbb{R}^{d_{L}\times d_0}$.
    \item The label noise $\xi_q \in \mathbb{R}^{d_{L}}$ is independent of $X$, with $\mathbb E[\xi_q]=0$, $X \perp \xi_q$ and covariance $\mathbb E[\xi_q \xi_q^\top] = \Sigma_q$.
\end{itemize}

The model error is denoted
\[
\Delta := M - W.
\]

For notational convenience, we define partial products of the student weights:
\[
W_{>l} := W_L W_{L-1}\cdots W_{l+1}, 
\qquad 
W_{<l} := W_{l-1}\cdots W_1,
\]
with the conventions $W_{>L}=I$ and $W_{<1}=I$. We also define
\[
A_l := W_{<l}, 
\qquad 
B_l := (W_{>l})^\top .
\]

\subsection*{Per--sample Gradient noise}

For one sample $(X,Y)$ with $Y=MX+\xi_q$ and $\Delta:=M-W$, the prediction error is:
\[
\varepsilon := Y - WX
\]
The per-sample squared loss is $\ell=\tfrac12\|\varepsilon\|^2$, and its gradient w.r.t.\ $W_l$ is
\[
g_l := \nabla_{W_l}\ell = - W_{>l}^{\top}\varepsilon X^{\top}W_{<l}^{\top}
\]
Vectorization and use of $\mathrm{vec}(uv^\top)=v\otimes u$ yields
\[
\mathrm{vec}(g_l)=-\mathrm{vec}\!\left((B_l\varepsilon)(A_lX)^\top\right)=-(A_lX)\otimes(B_l\varepsilon)
=-\big((A_lX)\otimes B_l\big)\varepsilon.
\]
Next, compute the population  gradient $g_l$. Since $g_l=-B_l(\varepsilon X^\top)A_l^\top$ and $A_l,B_l$ are deterministic given the current parameters,
\[
\mathbb E[g_l]=-B_l\,\mathbb E[\varepsilon X^\top]\,A_l^\top.
\]
Under whitened inputs $\mathbb E[XX^\top]=I_{d_0}$ this gives $\mathbb E[\varepsilon X^\top]=\Delta$ and therefore
\[
\mathbb E[g_l]= -\,B_l\Delta A_l^\top
\qquad\Longrightarrow\qquad
\mathbb E[\mathrm{vec}(g_l)]=-\mathrm{vec}(B_l\Delta A_l^\top)=-(A_l\otimes B_l)\,\mathrm{vec}(\Delta),
\]
where we used $\mathrm{vec}(AXB)=(B^\top\otimes A)\mathrm{vec}(X)$ with $A=B_l$, $X=\Delta$, $B=A_l^\top$. The per-sample gradient noise is
\[
\xi_l:=\mathbb E[\mathrm{vec}(g_l)] - \mathrm{vec}(g_l).
\]
Note the Kronecker identity
\[
(A_lX)\otimes B_l = (A_l\otimes B_l)(X\otimes I_{d_L}),
\]
and also $(X\otimes I)\varepsilon=\mathrm{vec}(\varepsilon X^\top)$. Hence,
\[
\mathrm{vec}(g_l)=-\big((A_lX)\otimes B_l\big)\varepsilon
=-(A_l\otimes B_l)(X\otimes I)\varepsilon
=-(A_l\otimes B_l)\,\mathrm{vec}(\varepsilon X^\top).
\]
Combining with the expression for $\mathbb E[\mathrm{vec}(g_l)]$ yields the exact factored form
\[
\boxed{
\xi_l=(A_l\otimes B_l)\Big(\mathrm{vec}(\varepsilon X^\top)-\mathrm{vec}(\Delta)\Big)
=(A_l\otimes B_l)\,\mathrm{vec}\!\left(\varepsilon X^\top - \Delta\right).
}
\]
Since $\mathbb E[\varepsilon X^\top]=\Delta$, we have that $\mathbb E[\xi_l]=0$.
In layerwise expression (not vectorized) the per-sample gradient noise can be written as:
\begin{equation*}
    \xi_l := W_{>l}^{\top}(\varepsilon X^{\top} - \Delta)W_{<l}^{\top}
\end{equation*}
\paragraph{Gradient-noise covariance.}
Recall the per-sample layerwise gradient noise for sample $(X, Y)$:
\[
\xi_l \;=\; (A_l\otimes B_l)\,\mathrm{vec}(\varepsilon X^\top-\Delta),\qquad
\varepsilon:=Y-WX=\Delta X+\xi_q,
\]
with $\Delta:=M-W$, whitened Gaussian inputs $X\sim\mathcal N(0,I_{d_0})$ (so $\E[XX^\top]=I_{d_0}$), and label noise $\xi_q\perp X$ with $\E[\xi_q]=0$ and $\E[\xi_q\xi_q^\top]=\Sigma_q$. Define the random matrix $S:=\varepsilon X^\top-\Delta\in\R^{d_{L}\times d_0}$ so that $\xi_l=(A_l\otimes B_l)\mathrm{vec}(S)$. Then for any pair of layers $l,m$:
\[
\Sigma_{lm}\;:=\;\E[\xi_l\xi_m^\top]
\;=\;(A_l\otimes B_l)\,\E[\mathrm{vec}(S)\mathrm{vec}(S)^\top]\,(A_m\otimes B_m)^\top.
\]
Using $\varepsilon=\Delta X+\xi_q$, expand
\[
S=(\Delta X+\xi_q)X^\top-\Delta=\Delta(XX^\top-I_{d_0})+\xi_qX^\top=:S_1+S_2.
\]
Since $\xi_q\perp X$ and $\E[\xi_q]=0$, the cross terms vanish:
\[
\E[\mathrm{vec}(S_1)\mathrm{vec}(S_2)^\top]=\E_X\!\Big[\mathrm{vec}(\Delta(XX^\top-I_{d_0}))\,\E_{\xi_q}[\mathrm{vec}(\xi_qX^\top)^\top\mid X]\Big]=0,
\]
and similarly $\E[\mathrm{vec}(S_2)\mathrm{vec}(S_1)^\top]=0$, hence $\E[\mathrm{vec}(S)\mathrm{vec}(S)^\top]=C_{\mathrm{data}}+C_{\mathrm{label}}$ with
\[
C_{\mathrm{data}}:=\E[\mathrm{vec}(S_1)\mathrm{vec}(S_1)^\top],\qquad
C_{\mathrm{label}}:=\E[\mathrm{vec}(S_2)\mathrm{vec}(S_2)^\top].
\]
For the data term, $\mathrm{vec}(S_1)=\mathrm{vec}(\Delta(XX^\top-I_{d_0}))=(I_{d_0}\otimes\Delta)\mathrm{vec}(XX^\top-I_{d_0})$, so
\[
C_{\mathrm{data}}=(I_{d_0}\otimes\Delta)\,\E[\mathrm{vec}(XX^\top-I_{d_0})\mathrm{vec}(XX^\top-I_{d_0})^\top]\,(I_{d_0}\otimes\Delta)^\top.
\]
Let $V:=\mathrm{vec}(XX^\top)=X\otimes X\in\R^{d_0^2}$. For index pairs $(i,j)$ and $(k,\ell)$, the $( (i,j),(k,\ell) )$ entry of $\E[VV^\top]$ is $\E[X_iX_jX_kX_\ell]$. By Wick/Isserlis for centered Gaussian vectors,
\[
\E[X_iX_jX_kX_\ell]=\E[X_iX_j]\E[X_kX_\ell]+\E[X_iX_k]\E[X_jX_\ell]+\E[X_iX_\ell]\E[X_jX_k].
\]
Since $\E[X_aX_b]=\delta_{ab}$ for $X\sim\mathcal N(0,I_{d_0})$, this becomes
\[
\E[X_iX_jX_kX_\ell]=\delta_{ij}\delta_{k\ell}+\delta_{ik}\delta_{j\ell}+\delta_{i\ell}\delta_{jk}.
\]
The three terms correspond respectively to $\mathrm{vec}(I_{d_0})\mathrm{vec}(I_{d_0})^\top$, the identity $I_{d_0^2}$, and the commutation matrix $C$ (defined by $C\mathrm{vec}(M)=\mathrm{vec}(M^\top)$). Thus
\[
\E[VV^\top]=I_{d_0^2}+C+\mathrm{vec}(I_{d_0})\mathrm{vec}(I_{d_0})^\top.
\]
Centering by $I_d$ yields
\[
\E[\mathrm{vec}(XX^\top-I_{d_0})\mathrm{vec}(XX^\top-I_{d_0})^\top]
=\E[(V-\mathrm{vec}(I_{d_0}))(V-\mathrm{vec}(I_{d_0}))^\top]
=\E[VV^\top]-\mathrm{vec}(I_{d_0})\mathrm{vec}(I_{d_0})^\top
=I_{d_0^2}+C,
\]
and therefore
\[
C_{\mathrm{data}}=(I_{d_0}\otimes\Delta)\,(I_{d_0^2}+C)\,(I_{d_0}\otimes\Delta)^\top.
\]
For the label term, $\mathrm{vec}(S_2)=\mathrm{vec}(\xi_qX^\top)=(X\otimes I_{d_{L}})\,\xi_q$, so using $\xi_q\perp X$,
\[
C_{\mathrm{label}}
=\E\!\big[(X\otimes I_{d_L})\xi_q\xi_q^\top(X\otimes I_{d_L})^\top\big]
=\E_X\!\big[(X\otimes I_{d_L})\Sigma_q(X\otimes I_{d_L})^\top\big]
=\E[XX^\top]\otimes\Sigma_q
=I_{d_0}\otimes\Sigma_q.
\]
Plugging $C_{\mathrm{data}}+C_{\mathrm{label}}$ into $\Sigma_{lm}$ and using $(A\otimes B)(I\otimes\Delta)=A\otimes(B\Delta)$ and $(A\otimes C)(B\otimes D)^\top=(AB^\top)\otimes(CD^\top)$ gives the final block covariance
\[
\boxed{
\Sigma_{lm}
=(A_l\otimes B_l\Delta)\,(I_{d_0^2}+C)\,(A_m\otimes B_m\Delta)^\top
\;+\;
(A_lA_m^\top)\otimes(B_l\Sigma_q B_m^\top).
}
\]
\section{Modewise diffusion on DLNs of depth L, proposition \ref{prop:mode-diffusion-main}}\label{app:diffusion-DLN}
Consider the input-output connectivity mode:
\begin{align*}
    w_{\alpha} = U^{\intercal\alpha}W_L...W_1V^{\alpha} 
\end{align*}
We want to empirically estimate the diffusion along mode $\alpha$. Let $\xi:=\text{vec}(\xi_l)$ be the stacked vector of gradient noise vectors. The noise covariance matrix of the whole DLN is given by:
\begin{align*}
    \Sigma = \mathbb{E}[\xi\xi^{\intercal}]
\end{align*}
The first-order perturbation of the mode amplitude is given by:
\begin{align*}
    \delta w_{\alpha} & = \sum_l U^{\intercal\alpha} W_{>l} d W_l W_{<l}V^{\alpha}\\
    W_{>l} & = W_L...W_{l+1}, \quad W_{<l} := W_{l-1}...W_1\\
    \delta w_{\alpha} & =  \text{Tr}(\delta w_{\alpha}) = \sum_l  \text{Tr}(A^{\intercal}_{l,\alpha}d W_l) = \sum_l \langle A_{l,\alpha},d W_l \rangle_F \quad \text{by the cyclicity of the trace}\\
    A_{l,\alpha} & := W_{>l}^{\intercal}U^{\alpha}V^{\intercal\alpha}W_{<l}^{\intercal}\\
\end{align*}
The diffusion of the amplitude of the mode $\alpha$ is therefore given by:
\begin{align*}
    D_{\alpha} & := \mathbb{E}[\delta w_{\alpha}^2] - \mathbb{E}[\delta w_{\alpha}]^2 \\
    & = a_{\alpha}^{\intercal}\text{Cov}(\text{vec}(dW_l)) a_{\alpha} \\
    D_{\alpha} & = \eta\sum_{lm} a_{l,\alpha}^{\intercal} \Sigma_{lm} a_{m,\alpha}\\
    a_{l,\alpha} & := \text{vec}(A_{l,\alpha}); \quad a_{\alpha} = \text{vec}(a_{l,\alpha})
\end{align*}
Note that we can also define the cross-mode diffusion:
\begin{align*}
    D_{\alpha\beta} & = \eta\sum_{lm} a_{l,\alpha}^{\intercal} \Sigma_{lm} a_{m,\beta} \\
\end{align*}
\section{Modewise state-dependent SDE over DLNs}\label{app:sde-ald-dlns}
\paragraph{Setup (stacked SDE approximating SGD).}
Let $L\ge 1$ be the number of layers. Fix time–independent input-output singular vectors $u_\alpha$ and $v_\alpha$ of appropriate dimensions, and define the mode amplitude:
\[
w_\alpha(t)\;:=\;u_\alpha^\top\big(W_{L,t}\cdots W_{1,t}\big)v_\alpha.
\]
For $l=1,\dots,L$ set the partial products
\[
W_{>l,t}:=W_{L,t}\cdots W_{l+1,t},
\qquad
W_{<l,t}:=W_{l-1,t}\cdots W_{1,t},
\]
with the convention that an empty product equals the identity.
\\
Stack all parameters
\begin{equation}
\theta_t:=\big(\mathrm{vec}\,W_{1,t};\,\dots;\,\mathrm{vec}\,W_{L,t}\big)\in\mathbb{R}^P,
\end{equation}
and let
\begin{equation}
    \Sigma(\theta)\;:=\;\mathbb{E}\!\big[\xi(\theta,Z)\,\xi(\theta,Z)^\top\mid \theta\big]
\end{equation}
be the \emph{conditional} covariance of the stacked one–step gradient noise $\xi$ induced by a minibatch $Z$. Write the block decomposition
$\Sigma=\big[\Sigma_{lm}\big]_{l,m=1}^L$ with $\Sigma_{lm}\in\mathbb{R}^{p_l\times p_m}$ and $p_l=d_ld_{l-1}$.
\paragraph{Block covariance across layers.}
For a minibatch $Z$ and stacked parameters $\theta=(\mathrm{vec}\,W_1;\dots;\mathrm{vec}\,W_L)$, let
\[
g_l(\theta;Z)\in\mathbb{R}^{d_l \times d_{l-1}} ,\qquad
g_l(\theta):=\mathbb{E}_Z\big[g_l(\theta;Z)\big],
\]
be, respectively, the minibatch gradient estimator and its population (or dataset) expectation for layer $l$. Define the layerwise gradient-noise vectors
\[
\xi_l(\theta,Z)\;:=\;\mathrm{vec}\big(g_l(\theta) - g_l(\theta;Z)\big)\in\mathbb{R}^{p_l},
\qquad p_l:=d_ld_{l-1},
\]
and stack $\xi(\theta,Z):=(\xi_1;\dots;\xi_L)\in\mathbb{R}^{P}$ with $P=\sum_l p_l$. The conditional covariance of the stacked noise is
\[
\Sigma(\theta)\;:=\;\mathbb{E}_Z\!\big[\xi(\theta,Z)\,\xi(\theta,Z)^\top\big]
\;=\;\big[\Sigma_{lm}(\theta)\big]_{l,m=1}^L,
\]
with blocks
\begin{equation}\label{eq:blockSigma}
\boxed{\;
\Sigma_{lm}(\theta)
\;:=\;
\mathbb{E}_Z\!\big[\;\xi_l(\theta,Z)\,\xi_m(\theta,Z)^\top\;\big]
\;}
\end{equation}
\\
\\
Choose any measurable matrix square root $\sigma(\theta)$ with $\sigma(\theta)\sigma(\theta)^\top=\Sigma(\theta)$ and drive the \emph{stacked} SDE with a standard $P$–dimensional Brownian motion $B_t$:
\begin{equation}\label{eq:stackedSDE}
d\theta_t \;=\; -\,g(\theta_t)\,dt \;+\; \sqrt{\eta}\,\sigma(\theta_t)\,dB_t,
\qquad \eta>0.
\end{equation}
This yields the quadratic covariation
\begin{equation}\label{eq:thetaBracket}
d[\theta]_t \;=\; \eta\,\Sigma(\theta_t)\,dt
\quad\Longleftrightarrow\quad
d\big[\mathrm{vec}\,W^l,\mathrm{vec}\,W^m\big]_t \;=\; \eta\,\Sigma_{lm}(\theta_t)\,dt .
\end{equation}

\paragraph{Quadratic covariation.}
For column vector semimartingales $X_t, Y_t$ with continuous paths, the \emph{quadratic covariation} (or bracket) is the matrix–valued, pathwise limit
\[
[X,Y]_t \;:=\; \lim_{|\Pi|\to 0} \sum_k \big(X_{t_{k+1}}-X_{t_k}\big)\big(Y_{t_{k+1}}-Y_{t_k}\big)^\top,
\]
taken along partitions $\Pi$ of $[0,t]$. For continuous It\^o processes, $[X,Y]_t$ equals the  bracket $\langle X,Y\rangle_t$, and if
$dX_t=H_X(t)\,dB_t$ and $dY_t=H_Y(t)\,dB_t$ are written against a  Brownian $B_t$, then
\[
d[X,Y]_t \;=\; H_X(t)\,H_Y(t)^\top\,dt.
\]
Applying this to \eqref{eq:stackedSDE} gives \eqref{eq:thetaBracket}.

\paragraph{First and second derivatives of the mode amplitude.}
Consider the scalar multilinear map
\[
f(W_1,\dots,W_L) \;=\; u_\alpha^\top W_L\cdots W_1 v_\alpha .
\]
Its derivative with respect to $W^l$ is the matrix
\begin{equation}\label{eq:Al}
A_{l,\alpha}(t) \;:=\; \partial_{W^l} f(W_{1,t},\dots,W_{L,t})
\;=\; (W_{>l,t}^\top u_\alpha)\,(W_{<l,t} v_\alpha)^\top ,
\end{equation}
so that for any perturbation $H$ one has $\mathrm{D}_{W_l} f[H]=\langle A_{l,\alpha}(t),H\rangle_F$ with
$\langle X,Y\rangle_F=\mathrm{Tr}(X^\top Y)$.
Because $f$ is linear in each argument, the \emph{diagonal} second derivatives vanish:
$\partial^2_{W_l,W_l} f\equiv 0$ for every $l$.
The \emph{mixed} derivatives $\partial^2_{W_l, W_m} f$ ($l\neq m$) are nonzero and are most cleanly described by their bilinear actions. Writing $H_l$ and $H_m$ for test directions,
\begin{equation}\label{eq:mixedHess-bilinear}
\mathrm{D}^2_{W_l,W_m} f[H_l,H_m] \;=\;
\begin{cases}
u_\alpha^\top\,W_{>l,t}\,H_l\,\big(W_{l-1,t}\cdots W_{m+1,t}\big)\,H_m\,W_{<m,t}\,v_\alpha, & l>m,\\[2pt]
u_\alpha^\top\,W_{>m,t}\,H_m\,\big(W_{m-1,t}\cdots W_{l+1,t}\big)\,H_l\,W_{<l,t}\,v_\alpha, & m>l.
\end{cases}
\end{equation}
Equivalently, in the stacked, vectorized coordinates $\theta=(\mathrm{vec}\,W_1;\dots;\mathrm{vec}\,W_L)$, let
\[
a_{l,\alpha}(t):=\mathrm{vec}\,A_{l,\alpha}(t),\qquad
a_\alpha(t):=\big(a_{1,\alpha}(t);\dots;a_{L,\alpha}(t)\big)\in\mathbb{R}^P,
\]
and for $l>m$, using the identification $\mathrm{D}^2_{W^l,W^m} f[H_l,H_m] = \text{vec}(H_l)^{\top}\partial_{\text{vec}(W_l),\text{vec}(W_m)}w_{\alpha}\text{vec}(H_m)$ we have: $$\nabla^2_{l,m} w_\alpha(\theta_t) = (W_{l-1:m+1} \otimes W_{>l}^{\top}u_\alpha)(v_{\alpha}^{\top}W_{<m}^{\top}\otimes I) $$ the full $p_l\times p_m$ Hessian of the mode amplitude. Then the diagonal blocks of $\nabla^2 w_\alpha$ are zero, while the off–diagonal blocks encode the bilinear forms in \eqref{eq:mixedHess-bilinear}.

\paragraph{It\^o differential of the mode amplitude.}
Applying the multivariate It\^o formula to $w_\alpha(t)=f(W_{1,t},\dots,W_{L,t})$ under the stacked SDE \eqref{eq:stackedSDE} yields
\begin{equation}\label{eq:ito-walpha}
dw_\alpha(t)
\,=\, \sum_{l=1}^L \big\langle A_{l,\alpha}(t),\, dW_{l,t}\big\rangle
\;+\;\frac12\, d[\theta]_t : \nabla^2 w_\alpha(\theta_t),
\end{equation}
where $A_{l,\alpha}$ is given by \eqref{eq:Al}, “$:$” denotes the Frobenius contraction in the stacked coordinates, and the second term collects all mixed second–order contributions. Substituting \eqref{eq:stackedSDE} and \eqref{eq:thetaBracket} into \eqref{eq:ito-walpha} gives
\begin{equation}\label{eq:ito-expanded}
dw_\alpha(t)
\,=\, -\sum_{l=1}^L \langle A_{l,\alpha}(t),\, g_l(\theta_t)\rangle\,dt
\;+\;\sqrt{\eta}\,a_\alpha(t)^\top \sigma(\theta_t)\,dB_t
\;+\;\frac{\eta}{2}\,\mathrm{tr}\!\big(\Sigma(\theta_t)\,\nabla^2 w_\alpha(\theta_t)\big)\,dt.
\end{equation}
The last term is the It\^o \emph{drift correction}. It vanishes when $\Sigma(\theta_t)$ is block–diagonal (in which case $d[\mathrm{vec}\,W^l,\mathrm{vec}\,W^m]_t\equiv 0$ for $l\neq m$); off-diagonal blocks $\Sigma_{lm}$ are generally nonzero and \eqref{eq:mixedHess-bilinear} weight their contribution to the It\^o induced drift.

\paragraph{Diffusion (variance rate) of the mode amplitude.}
Write $dw_\alpha=\mu_\alpha(\theta_t)\,dt+dM_{\alpha,t}$, where the martingale part is
\[
dM_{\alpha,t}\;=\;\sqrt{\eta}\,a_\alpha(t)^\top \sigma(\theta_t)\,dB_t .
\]
Conditioning on the natural filtration $\mathcal{F}_t$ and using that $a_\alpha(t)$ and $\sigma(\theta_t)$ are $\mathcal{F}_t$–measurable, one obtains
\[
\mathbb{E}\!\big[(dM_{\alpha,t})^2\mid \mathcal{F}_t\big]
= \eta\,a_\alpha(t)^\top \sigma(\theta_t)\,
\mathbb{E}[\,dB_t\,dB_t^\top\mid\mathcal{F}_t]\,\sigma(\theta_t)^\top a_\alpha(t)
= \eta\,a_\alpha(t)^\top \Sigma(\theta_t)\,a_\alpha(t)\;dt .
\]
Therefore, the instantaneous variance rate (diffusion coefficient) is
\begin{equation}\label{eq:Dalp-full}
\boxed{
D_\alpha(\theta_t)
\;:=\;
\frac{1}{dt}\,
\mathbb{E}\!\big[(dw_\alpha-\mathbb{E}[dw_\alpha\mid\mathcal{F}_t])^2\mid \mathcal{F}_t\big]
\;=\;
\eta\,a_\alpha(t)^\top \Sigma(\theta_t)\,a_\alpha(t)
\;=\;
\eta\sum_{l,m=1}^L a_{l,\alpha}(t)^\top \Sigma_{lm}(\theta_t)\,a_{m,\alpha}(t).}
\end{equation}
Equivalently, there exists a one-dimensional Brownian motion $\beta_{\alpha,t}$ such that
\[
dw_\alpha(t)\;=\;\mu_\alpha(\theta_t)\,dt \;+\; \sqrt{D_\alpha(\theta_t)}\,d\beta_{\alpha,t},
\]
with $D_\alpha$ given by \eqref{eq:Dalp-full}. Analogously, the cross–mode diffusion is, for any $\alpha,\beta$,
\begin{equation}
\boxed{
D_{\alpha,\beta}:=\frac{1}{dt}\,\mathbb{E}\!\big[dM_{\alpha,t}\,dM_{\beta,t}\mid\mathcal{F}_t\big]
\;=\; \eta\,a_\alpha(t)^\top \Sigma(\theta_t)\,a_\beta(t)
\;=\; \eta\sum_{l,m=1}^L a_{l,\alpha}(t)^\top \Sigma_{lm}(\theta_t)\,a_{m,\beta}(t)}
\end{equation}

\paragraph{Remarks.}
The cross–layer diffusion arises from the off–diagonal blocks $\Sigma_{lm}(\theta_t)$ induced by the shared minibatch. The drift correction in \eqref{eq:ito-expanded} involves only the \emph{mixed} second derivatives of $w_\alpha$ via \eqref{eq:mixedHess-bilinear}; the diagonal second derivatives vanish because mode amplitude $w_{\alpha}$ is linear in each $W_l$ separately.
\section{Scalar modewise SDE for aligned mode under balanced conditions; proof of proposition~\ref{prop:modewise-SDE-balanced}}
\label{app:proof-of-modewise-SDE-balanced}
\paragraph{Set-up and notation.}
On the aligned and balanced manifold, the mode amplitude and diagonal entries of the weight matrices $w_{l,\alpha}$ satisfy:
\[
w_\alpha \;=\; u_\alpha^\top W_L\cdots W_1 v_\alpha,
\qquad
w_{l,\alpha}:= w_\alpha^{\frac{1}{L}}
\]
For each layer $l$, define the sensitivities
\[
A_{l,\alpha}
:=\partial_{W^l} w_\alpha
=(W_{>l}^{\!\top}u_\alpha)\,(W_{<l}v_\alpha)^\top,
\qquad
a_{l,\alpha}:=\mathrm{vec}(A_{l,\alpha})
=(W_{<l}v_\alpha)\otimes(W_{>l}^{\!\top}u_\alpha),
\]
with $W_{<l}:=W_{l-1}\cdots W_{1}$. On the aligned manifold (no cross-modes),
\[
\|A_{l,\alpha}\|_F^2
=\|W_{<l}v_\alpha\|^2\,\|W_{>l}^{\!\top}u_\alpha\|^2
=\prod_{j\neq l} w_{j,\alpha}^{2}.
\]

\paragraph{Stacked SDE and It\^o formula.}
Let the stacked parameter SDE be
\[
d\theta_t=-g(\theta_t)\,dt+\sqrt{\eta}\,\sigma(\theta_t)\,dB_t,
\quad \sigma\sigma^\top=\Sigma,
\]
with $\theta=(\mathrm{vec}\,W_1;\dots;\mathrm{vec}\,W_L)$ and block covariance $\Sigma=\{\Sigma_{lm}\}_{l,m=1}^L$.
By multivariate It\^o for the scalar $w_\alpha(\theta)$,
\begin{equation}\label{eq:ito-master}
dw_\alpha
=
\underbrace{\Big(-\sum_{l=1}^L\langle A_{l,\alpha},\,g_l\rangle\Big)}_{\mu_\alpha^{\rm grad}}
\,dt
\;+\;
\underbrace{\frac{\eta}{2}\,\mathrm{tr}\!\big(\Sigma\,\nabla^2 w_\alpha\big)}_{\mu_\alpha^{\rm Ito}}
\,dt
\;+\;
\underbrace{\sqrt{\eta}\,a_\alpha^\top \sigma\,dB_t}_{\text{diffusion}},
\qquad
a_\alpha:=(a_{1,\alpha};\dots;a_{L,\alpha}).
\end{equation}

\subsection*{1.\; Gradient drift $\mu_\alpha^{\rm grad}$ (aligned $\Rightarrow$ modewise GF)}
Population gradient blocks for squared loss with whitened inputs read $g_l=W_{>l}^{\!\top}(W-M)W_{<l}^{\!\top}$. Under alignment,
\[
(W-M)\Big|_{\alpha}=(w_\alpha-s_\alpha)\,u_\alpha v_\alpha^\top
\quad\Longrightarrow\quad
g_l\Big|_{\alpha}=(w_\alpha-s_\alpha)\,A_{l,\alpha}.
\]
Hence
\[
\mu_\alpha^{\rm grad}
=-\sum_{l=1}^L\langle A_{l,\alpha},g_l\rangle
=(s_\alpha-w_\alpha)\sum_{l=1}^L\|A_{l,\alpha}\|_F^2
=(s_\alpha-w_\alpha)\sum_{l=1}^L\prod_{j\neq l} w_{j,\alpha}^{2}.
\]
Imposing balance $w_{1,\alpha}=\cdots=w_{L,\alpha}=w_\alpha^{1/L}$ gives
\[
\mu_\alpha^{\rm grad}(w_\alpha)=(s_\alpha-w_\alpha)\,L\,w_\alpha^{\frac{2(L-1)}{L}},
\]
which is \eqref{eq:grad-drift}.

\subsection*{2.\; It\^o drift $\mu_\alpha^{\rm Ito}$ (off-diagonal Hessian $\times$ covariance)}
Because $w_\alpha$ is multilinear in the $\{W_l\}$, diagonal Hessian blocks vanish, and only off-diagonal blocks contribute:
\[
\mu_\alpha^{\rm Ito}
=\frac{\eta}{2}\sum_{l\neq m}\!\langle \Sigma_{lm},\,H_{lm}[w_\alpha]\rangle_F.
\]
The mixed Hessian block $H_{lm}$ is the bilinear form (for $l>m$; the other case is symmetric)
\[
D^2_{W^l,W^m} w_\alpha[H_l,H_m]
=
u_\alpha^\top \,W_{>l}\,H_l\,
(W_{l-1}\cdots W_{m+1})\,H_m\,W_{<m}\,v_\alpha .
\]
The SGD noise covariance (whitened inputs, label noise independent of $X$) decomposes as
\[
\Sigma_{lm}
=(W_{<l}\otimes B_l\Delta)(I_{d_0^2} + C )(W_{<m}\otimes B_m\Delta)^\top \;+\; (W_{<l}W_{<m}^\top)\otimes(B_l\Sigma_q B_m^\top),
\]
with $B_l=W_{>l}^\top$, $\Delta:=M-W$ and $C$ the commutation matrix.

\paragraph{Evaluate on the aligned manifold.}
On alignment, $\Delta|_\alpha=(s_\alpha-w_\alpha)u_\alpha v_\alpha^\top$. Using
$\mathrm{vec}(uv^\top)=v\otimes u$, $C(x\otimes y)=y\otimes x$, and
$(x\otimes y)^\top(A\otimes B)(u\otimes v)=(x^\top Au)(y^\top Bv)$, one finds the two contractions:

\emph{(i) Data–mismatch term.}
\[
\sum_{l\neq m}\!\big\langle (A_l\!\otimes\!B_l\Delta)(I+C)(A_m\!\otimes\!B_m\Delta)^\top,\;H_{lm}\big\rangle_F
=2\,(s_\alpha-w_\alpha)\sum_{l\neq m}\prod_{j\neq l,m}w_{j,\alpha}^{2}.
\]

\emph{(ii) Label–noise term.}
\[
\sum_{l\neq m}\!\big\langle (A_lA_m^\top)\!\otimes\!(B_l\Sigma_q B_m^\top),\;H_{lm}\big\rangle_F
=\sum_{l\neq m}\!\Big(\prod_{j\neq l,m}w_{j,\alpha}^{2}\Big)\;
\big\langle u_\alpha u_\alpha^\top,\;B_l\Sigma_q B_m^\top\big\rangle.
\]

\paragraph{Impose balance.}
Since $\sum_{l\neq m}\prod_{j\neq l,m}w_{j,\alpha}^{2}=L(L-1)\,w_\alpha^{\frac{2(L-2)}{L}}$, the It\^o drift becomes
\[
\mu_\alpha^{\rm Ito}(w_\alpha)
=\frac{\eta}{2}\,\Big[
2\,(s_\alpha-w_\alpha)\,L(L-1)\,w_\alpha^{\frac{2(L-2)}{L}}
\;+\;L(L-1)\,w_\alpha^{\frac{2(L-2)}{L}}
\,\Gamma_\alpha\Big],
\]
i.e.
\[
\mu_\alpha^{\rm Ito}(w_\alpha)
=\eta\,(s_\alpha-w_\alpha)\,L(L-1)\,w_\alpha^{\frac{2(L-2)}{L}}
\;+\;\frac{\eta}{2}\,\,L(L-1)\,w_\alpha^{\frac{2(L-2)}{L}}\;\Gamma_\alpha,
\]
which is \eqref{eq:ito-drift}. Here
\[
\Gamma_\alpha:=\frac{1}{L(L-1)}\sum_{l\neq m}\big\langle u_\alpha u_\alpha^\top,\;B_l\Sigma_q B_m^\top\big\rangle.
\]

\subsection*{3.\; Diffusion coefficient $D_\alpha$ (mode-diagonal)}
The scalar diffusion along mode $\alpha$ is
\[
D_\alpha=\eta\,a_\alpha^\top \Sigma\,a_\alpha
=\eta\sum_{l,m} a_{l,\alpha}^\top \Sigma_{lm}\,a_{m,\alpha}.
\]
Orthogonality of different modes under alignment implies $D_{\alpha\beta}=0$ for $\alpha\neq\beta$ (no cross-mode diffusion).

\paragraph{Data–mismatch part.}
A direct application of the vector identities yields, for each $(l,m)$,
\[
a_{l,\alpha}^\top (A_l\!\otimes\!B_l\Delta)(I_{d_0^2} + C)(A_m\!\otimes\!B_m\Delta)^\top a_{m,\alpha}
=2\,(s_\alpha-w_\alpha)^2
\Big(\prod_{j\neq l}w_{j,\alpha}^{2}\Big)\Big(\prod_{j\neq m}w_{j,\alpha}^{2}\Big).
\]
Summing and multiplying by $\eta$ gives
\[
D_\alpha^{\rm data}
=2\,\eta\,(s_\alpha-w_\alpha)^2\Bigg[\sum_{l=1}^L \prod_{j\neq l}w_{j,\alpha}^{2}\Bigg]^2.
\]

\paragraph{Label–noise part.}
Similarly,
\[
a_{l,\alpha}^\top\big((A_lA_m^\top)\!\otimes\!(B_l\Sigma_q B_m^\top)\big)a_{m,\alpha}
=\Big(\prod_{j\neq l}w_{j,\alpha}^{2}\Big)\Big(\prod_{j\neq m}w_{j,\alpha}^{2}\Big)\,\Gamma_{lm,\alpha},
\quad
\Gamma_{lm,\alpha}:=\big\langle u_\alpha u_\alpha^\top,\;B_l\Sigma_q B_m^\top\big\rangle,
\]
hence
\[
D_\alpha^{\rm label}
=\eta\sum_{l,m}\Big(\prod_{j\neq l}w_{j,\alpha}^{2}\Big)\Big(\prod_{j\neq m}w_{j,\alpha}^{2}\Big)\Gamma_{lm,\alpha}.
\]

\paragraph{Impose balance.}
Since $\sum_{l}\prod_{j\neq l}w_{j,\alpha}^{2}=L\,w_\alpha^{\frac{2(L-1)}{L}}$, we obtain
\[
D_\alpha^{\rm data}
=2\,\eta\,\,L^{2}(s_\alpha-w_\alpha)^2\,w_\alpha^{\frac{4(L-1)}{L}},
\qquad
D_\alpha^{\rm label}
=\eta\,\,L^{2}\,\overline{\Gamma}_\alpha\,w_\alpha^{\frac{4(L-1)}{L}},
\]
where $\displaystyle \overline{\Gamma}_\alpha:=\frac{1}{L^{2}}\sum_{l,m}\Gamma_{lm,\alpha}$. Therefore
\[
D_\alpha(w_\alpha)
=\eta\,\,L^{2}\Big(2\,(s_\alpha-w_\alpha)^2+\overline{\Gamma}_\alpha\Big)\,
w_\alpha^{\frac{4(L-1)}{L}},
\]
which is \eqref{eq:diffusion}. In the isotropic case $\Sigma_q=\sigma_q^2 I$ one has $\Gamma_{lm,\alpha}=\sigma_q^2$ and hence
$\overline{\Gamma}_\alpha=\sigma_q^2$.

\section{Derivation of modewise stationary law under detailed balance, proposition \ref{prop:modewise-stationary}}\label{sec:proof stationary-law}
The Fokker--Planck equation for the modewise density $p_\alpha(w,t)$ is
\begin{equation}\label{eq:fp}
\partial_t p_\alpha(w,t)
= -\partial_w\!\big(\mu_\alpha(w)\,p_\alpha(w,t)\big)
+ \tfrac12\,\partial_w^2\!\big(D_\alpha(w)\,p_\alpha(w,t)\big).
\end{equation}
Define the probability current $J(w,t):=\mu_\alpha(w)\,p_\alpha - \tfrac12\,\partial_w\!\big(D_\alpha p_\alpha\big)$. Under detailed balance (zero current in steady state), $J= 0$ and \eqref{eq:fp} reduces at stationarity to the first-order ODE
\[
\tfrac12\,(D_\alpha p_\alpha^\star)' - \mu_\alpha\,p_\alpha^\star \;=\; 0.
\]
Divide by $D_\alpha>0$ and write it in linear form $(\log p_\alpha^\star)'+(\log D_\alpha)' = \tfrac{2\mu_\alpha}{D_\alpha}$. Integrating yields \eqref{eq:general-stationary} (up to a normalization constant).

For (i) $\sigma_q=0$: as $w\uparrow s_\alpha$, one has
\[
\mu_\alpha(w)=L\,w^{a}(s_\alpha-w)+O(\eta),\qquad
D_\alpha(w)=2\beta\,L^2\,(s_\alpha-w)^2 w^{2a}.
\]
Hence $\tfrac{2\mu_\alpha}{D_\alpha}\sim \tfrac{c}{(s_\alpha-w)}$ with $c>0$, so the exponential in \eqref{eq:general-stationary} diverges like $(s_\alpha-w)^{-\gamma}$ with $\gamma>0$ and the prefactor $1/D_\alpha$ contributes another $(s_\alpha-w)^{-2}$. The resulting singularity is non-integrable at $s_\alpha$, which becomes an absorbing state forcing the stationary measure to collapse to $\delta(w-s_\alpha)$.

For (ii) $\sigma_q>0$: evaluate drift, slope, and diffusion at $w=s_\alpha$.
\begin{align*}
\mu_\alpha(s_\alpha)
&=\mu_\alpha^{\rm grad}(s_\alpha)+\mu_\alpha^{\rm Ito}(s_\alpha)
= 0 \;+\; \frac{\eta}{2B}\,L(L-1)\,s_\alpha^{b}\,\sigma_q,\\
\mu_\alpha'(s_\alpha)
&= \frac{d}{dw}\Big[L\,w^{a}(s_\alpha-w)\Big]_{w=s_\alpha} + O(\eta)
= -L\,s_\alpha^{a} + O(\eta),\\
D_\alpha(s_\alpha)
&= \beta\,L^2\,\sigma_q^2\,s_\alpha^{2a}.
\end{align*}
A first-order zero of $\mu_\alpha(w)$ is obtained by linearizing: $0=\mu_\alpha(s_\alpha)+\mu_\alpha'(s_\alpha)(w_\alpha^\star-s_\alpha)+O(\eta^2)$, hence
\[
w_\alpha^\star-s_\alpha
= -\frac{\mu_\alpha(s_\alpha)}{\mu_\alpha'(s_\alpha)} + O(\eta^2)
\]
This gives
\[
w_\alpha^\star-s_\alpha
= \frac{\eta}{2B}\,(L-1)\,s_\alpha^{-2/L}\,\sigma_q + O(\eta^2).
\]
\paragraph{Variance and local linearization.}
To compute the variance of the stationary distribution, we linearize the scalar SDE in a small neighborhood of the stable fixed point $w_\alpha^\star$, defined by $\mu_\alpha(w_\alpha^\star)=0$.  
Setting $x_t := w_t - w_\alpha^\star$, the dynamics expand as
\[
dx_t
= \mu_\alpha'(w_\alpha^\star)\,x_t\,dt
+ \sqrt{D_\alpha(w_\alpha^\star)}\,dB_{\alpha,t}
+ O(\eta),
\]
where higher-order terms such as $x_t^2$ or $x_t D_\alpha'(w_\alpha^\star)$ are $O(\eta)$ smaller than the leading terms because $x_t=O(\sqrt{\eta})$ in the stationary regime (since $D_\alpha=O(\eta)$).  
Neglecting these subleading corrections yields an Ornstein--Uhlenbeck (OU) approximation.
\[
dx_t = \mu_\alpha'(w_\alpha^\star)\,x_t\,dt + \sigma\,dB_{\alpha,t},
\qquad
\sigma^2 := D_\alpha(w_\alpha^\star).
\]
For an OU process, It\^o's formula applied to $x_t^2$ gives
\[
d(x_t^2) = 2 \mu_\alpha'(w_\alpha^\star)\,x_t^2\,dt + \sigma^2\,dt + 2\sigma\,x_t\,dB_{\alpha,t},
\]
and taking expectations at stationarity ($d\,\mathbb{E}[x_t^2]/dt=0$) yields
\[
0 = 2\mu_\alpha'(w_\alpha^\star),\operatorname{Var}(x_t) + \sigma^2
\quad\Longrightarrow\quad
\operatorname{Var}(x_t)
= \frac{\sigma^2}{-2\mu_\alpha'(w_\alpha^\star)}
= \frac{D_\alpha(w_\alpha^\star)}{-2\,\mu_\alpha'(w_\alpha^\star)}.
\]
Since $w_\alpha^\star = s_\alpha + O(\eta)$ and $D_\alpha'(s_\alpha),\mu_\alpha''(s_\alpha)=O(\eta)$,
evaluating at $s_\alpha$ instead of $w_\alpha^\star$ introduces only $O(\eta^2)$ corrections, so
\[
\operatorname{Var}_{p_\alpha^\star}(w_\alpha)
= \frac{D_\alpha(s_\alpha)}{-2\,\mu_\alpha'(s_\alpha)} + O(\eta^2)
= \frac{\beta L^2\sigma_q^2 s_\alpha^{2a}}{2L s_\alpha^{a}} + O(\eta^2)
= \frac{\beta L}{2}\,\sigma_q^2\,s_\alpha^{a} + O(\eta^2),
\]
with $a=\tfrac{2(L-1)}{L}$. \hfill$\square$

\section{Offline, finite-dataset case}
\label{app:finite-dataset-case}

This appendix provides a sketch of how the theory can be adapted to the finite-dataset case, as opposed to the online learning case assumed elsewhere.

\subsection{Finite population correction}

The batch is sampled without replacement from a finite dataset $\mathcal{D}_N=(x_i, y_i)_{i=1}^N$, so the covariance matrix for batch sizes $1<b$ has a finite-population correction rather than a $\frac{1}{b}$ factor.

By expanding the expectation over batches $B\subset \mathcal{D}_N$ with $|B|=b$, we can show that the batch gradient noise covariance matrix relates to the empirical one-sample gradient noise covariance matrix as follows:
\begin{equation}
    \Sigma_b = \frac{1}{b^2}\sum_{i,j=1}^N g(x_i)g(x_j)\mathbb{E}_{B}[1_B(x_i,x_j)] - g_N g_N^{\intercal}
    \label{batch-noise covariance}
\end{equation}
where $1_B(X_i,X_j)$ is the indicator function for samples $i$ and $j$ co-occuring in a batch. Sampling without replacement, the joint probability of $X_i$ and $X_j$ being in batch $B$ is given by a product of hypergeometric distributions:
\begin{equation*}
    p(X_i, X_j\in B) = p(X_i\in B|X_j\in B)P(X_j\in B) = \frac{\binom{B-2}{N-2}}{\binom{B-1}{N-1}}\frac{\binom{B-1}{N-1}}{\binom{B}{N}} = \frac{B-1}{N-1}\frac{B}{N}
\end{equation*} Plugging this joint probability in equation \ref{batch-noise covariance} we find:
\begin{equation*}
    \Sigma_b = \frac{N-B}{B(N-1)}\sum_{i\neq j}^N g_ig_j^{\intercal} + \frac{1}{B^2}\sum_{i}g_ig_i^{\intercal} - g_N g_N^{\intercal}
\end{equation*}
Furthermore, we have:
\begin{equation*}
    \sum_{i\neq j}g_ig_j^{\intercal} = N^2g_N g_N^{\intercal} -\sum_i g_ig_i^{\intercal}
\end{equation*}
This allows us to simplify the batch-noise covariance into the following equation:
\begin{equation*}
    \Sigma_b = \frac{N-B}{B(N-1)}\left[\frac{1}{N}\sum_{i=1}^{N}g_i(w)g_i(w)^{\intercal} - g(w)g(w) \right] = \frac{N-B}{B(N-1)}\Sigma
\label{batch_noise_cov_simplified}
\end{equation*}

\subsection{Expectations deviate from population mean}

The calculation of the covariance matrix takes expectations over the data distribution. In the finite dataset case, the covariance matrix therefore differ. As the dataset size $N$ tends to infinity, the statistics converge due to the law of large numbers.
\section{Experimental setup}
\label{app:experimental_setup}
\subsection{Learning task}

Unless noted otherwise, the learning task used in experiments is given by:
\begin{itemize}
    \item A teacher matrix $M$ with three non-zero singular values (correpsonding to modes to be learned). The singular values of the teacher matrix are in arithmetic progression: $1.0, 0.7, 0.4$.
    \item A dataset sampled from a stndard Gaussian distribution $X\sim\mathcal{N}(0 , I_{d_{in}})$. Unless otherwise specified, online learning is used, and data is sampled from the distribution independently at each parameter update step. The default data dimension is $d=12$.
    \item Labels $Y := MX + \xi_q$, where $M$ is the teacher matrix and $\xi_q$ is optional label noise.
    \item Mean-square-error loss function.
\end{itemize}

\subsection{Architecture and initialization}

By default, the architecture consists of square matrices $W\in \mathbb{R}^{12 \times 12}$ with variable depth (depth-two and depth-four networks are used most frequently, corresponding respectively to the setup of \cite{saxe2013exact} and a deeper network where calculating the gradient noise covariance matrix is still tractable).

We use a small initialization, so that training takes place in the rich regime. Specifically, each weight matrix $W\in\mathbb{R}^{d_1\times d_0}$ is initialized using i.i.d. samples from a Gaussian distribution with mean $0$ and variance $\text{min}\{d_0, d_1\}^{-\gamma}$ where $\gamma$ is a hyperparameter controlling initialization scale. Values of $\gamma>1$ correspond to the rich regime \citep{jacot2021saddle}, and by default we use $\gamma=3$ to be well within the regime.

(\autoref{fig:summary} uses $\gamma=2$ so that the times of mode learning are more concentrated and easier to visualize.)

\subsection{Training}

Several optimizers are used:

\begin{itemize}
    \item \textbf{Gradient descent}: a large fixed dataset is used in computing a gradient.
    \item \textbf{SGD} with batch size $b$: a subset of batch size $b$ is randomly selected from the data distribution at every iteration to compute the gradient (this is online SGD, which our theoretical results are about). 
    \item \textbf{Isotropic Langevin dynamics}: the gradient is computed as in gradient descent, and an isotropic Gaussian noise term $\xi \sim \mathcal{N}(\mathbf{0}, \eta \, \Delta t \,I)$ is added, where $\eta$ is the learning rate and $\Delta t$ is the discretization time step. This corresponds to an Euler-Maruyama discretization of a corresponding Langevin dynamics SDE.
    \item \textbf{Anisotropic, state-dependent noise}: the gradient is computed as in gradient descent, and noise sampled using the SGD noise covariance $\Sigma(\theta)$ is added. The covariance matrix is recomputed at each step. The noise is therefore both anisotropic (the covariance matrix is not identity) and nonhomogeneous (the covariance matrix varies along the trajectory). This corresponds to an Euler-Maruyama simulation of \autoref{eq:sgd-sde}.
\end{itemize}

The default learning rate used is $\eta=0.005$. SGD by default uses batch size 1. For the Euler-Maruyama simulations, we use $\Delta t=10^{-4} \ll \eta$ for accuracy.

\section{Testing assumptions}\label{app:testing-assumptions}

This appendix describes experiments that test how well the balance and alignment assumptions, \ref{assum:balance} and \ref{assum:align}, hold under standard online SGD training.

\begin{figure}
    \centering
    \begin{subfigure}{0.6\textwidth}
        \includegraphics[width=\linewidth]{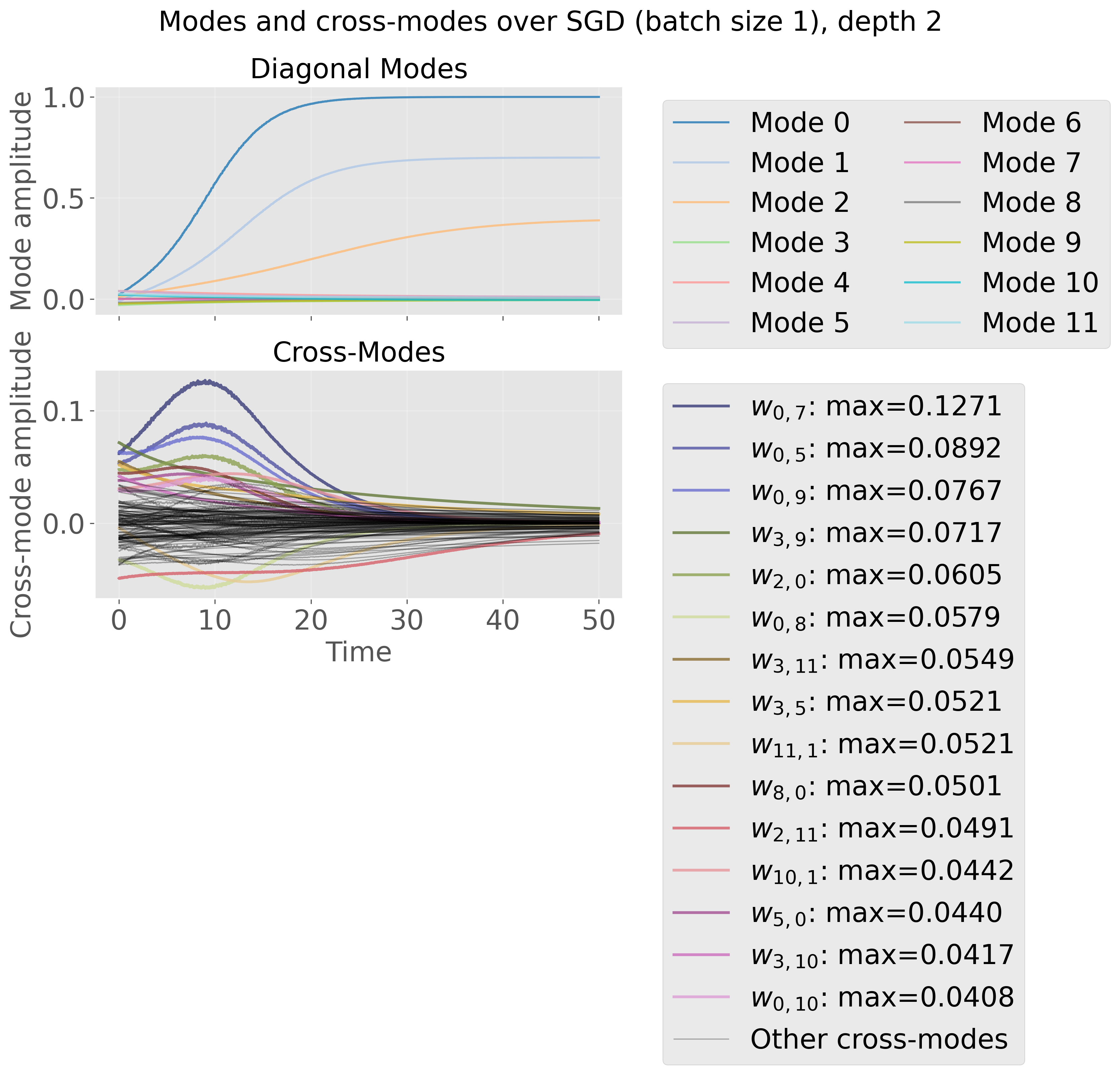}
        \caption{Depth 2}
    \end{subfigure}
    \begin{subfigure}{0.6\textwidth}
        \includegraphics[width=\linewidth]{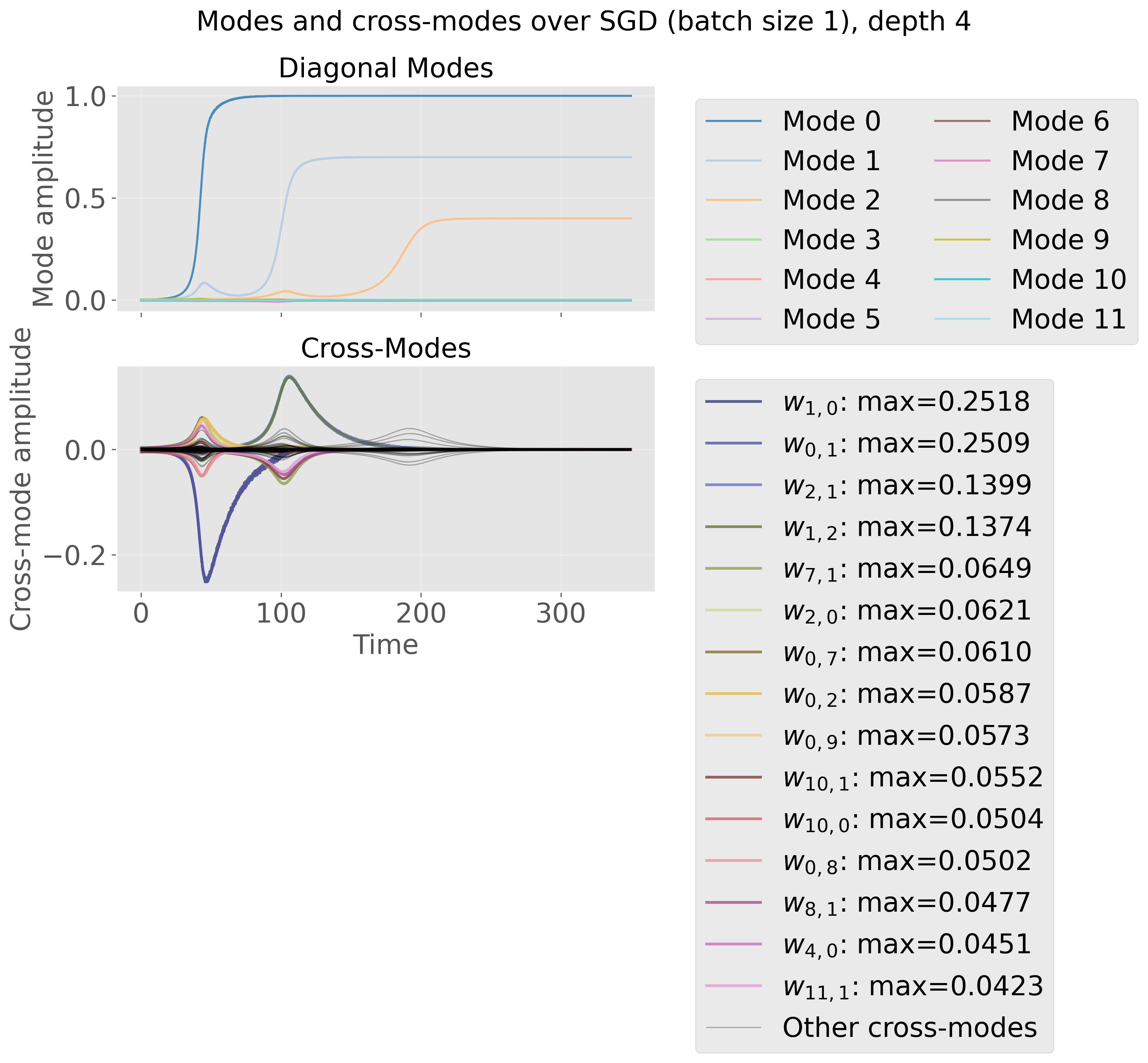}
        \caption{Depth 4}
    \end{subfigure}
    \caption{\textbf{Measuring magnitude of cross modes over SGD training}. For both a depth-2 and a depth-4 linear network, we observe that the majority of cross modes are small for most of training, except for a small number of cross modes which peak at times corresponding to modes being learned.}
    \label{fig:cross_modes}
\end{figure}

\subsection{Alignment}

\autoref{fig:cross_modes} examines the assumption that cross modes can be neglected. We observe that for most of training, cross modes are negligible, except for a few that peak during the intervals when modes are learned. The fact that the 0-1 cross mode has the largest peak is perhaps related to the fact that mode 1 has a visible increase while mode 0 is being learned.

\begin{figure}
    \centering
    \begin{subfigure}{0.48\textwidth}
        \includegraphics[width=\linewidth]{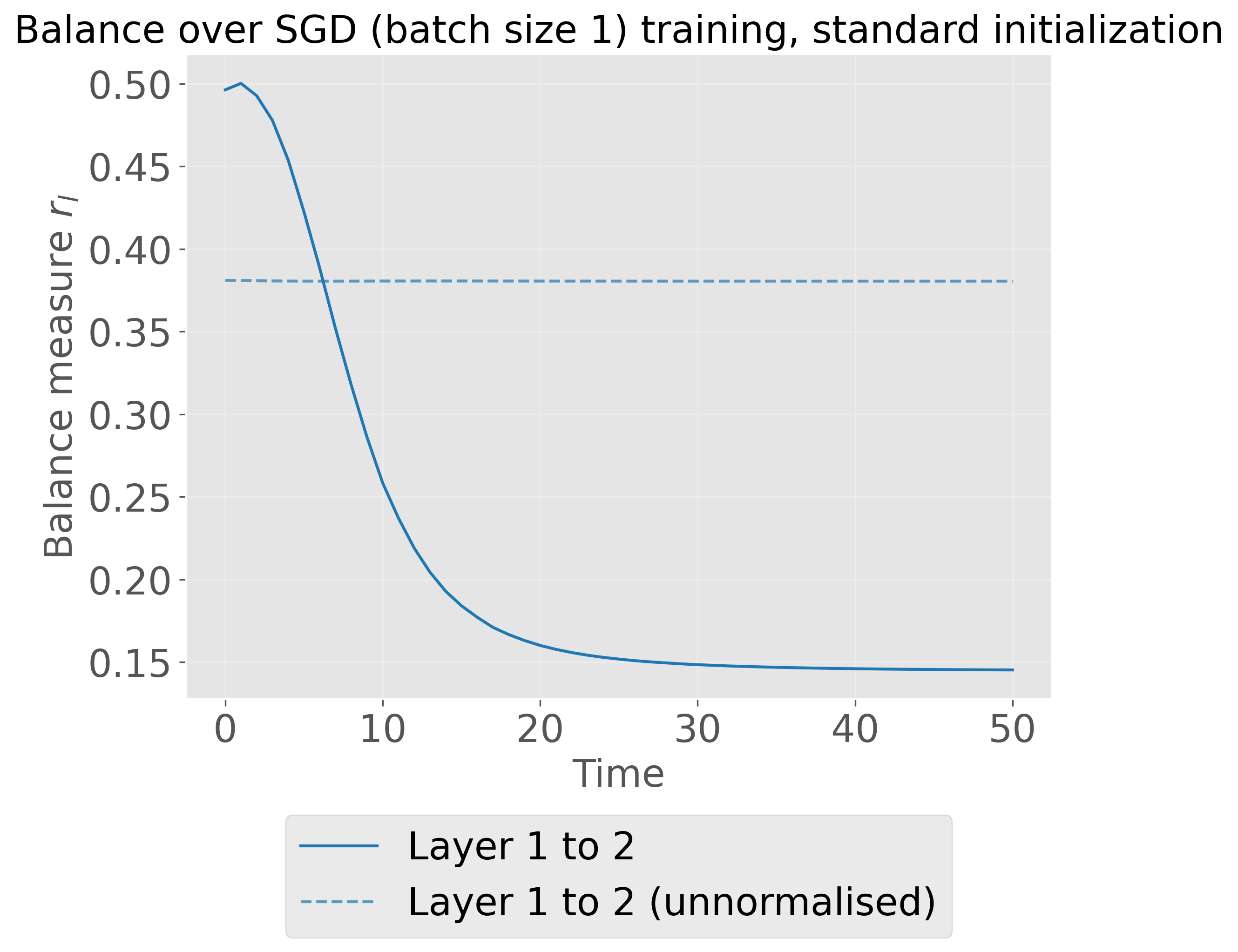}
        \caption{Standard initialization, depth 2}
    \end{subfigure}
    \begin{subfigure}{0.48\textwidth}
        \includegraphics[width=\linewidth]{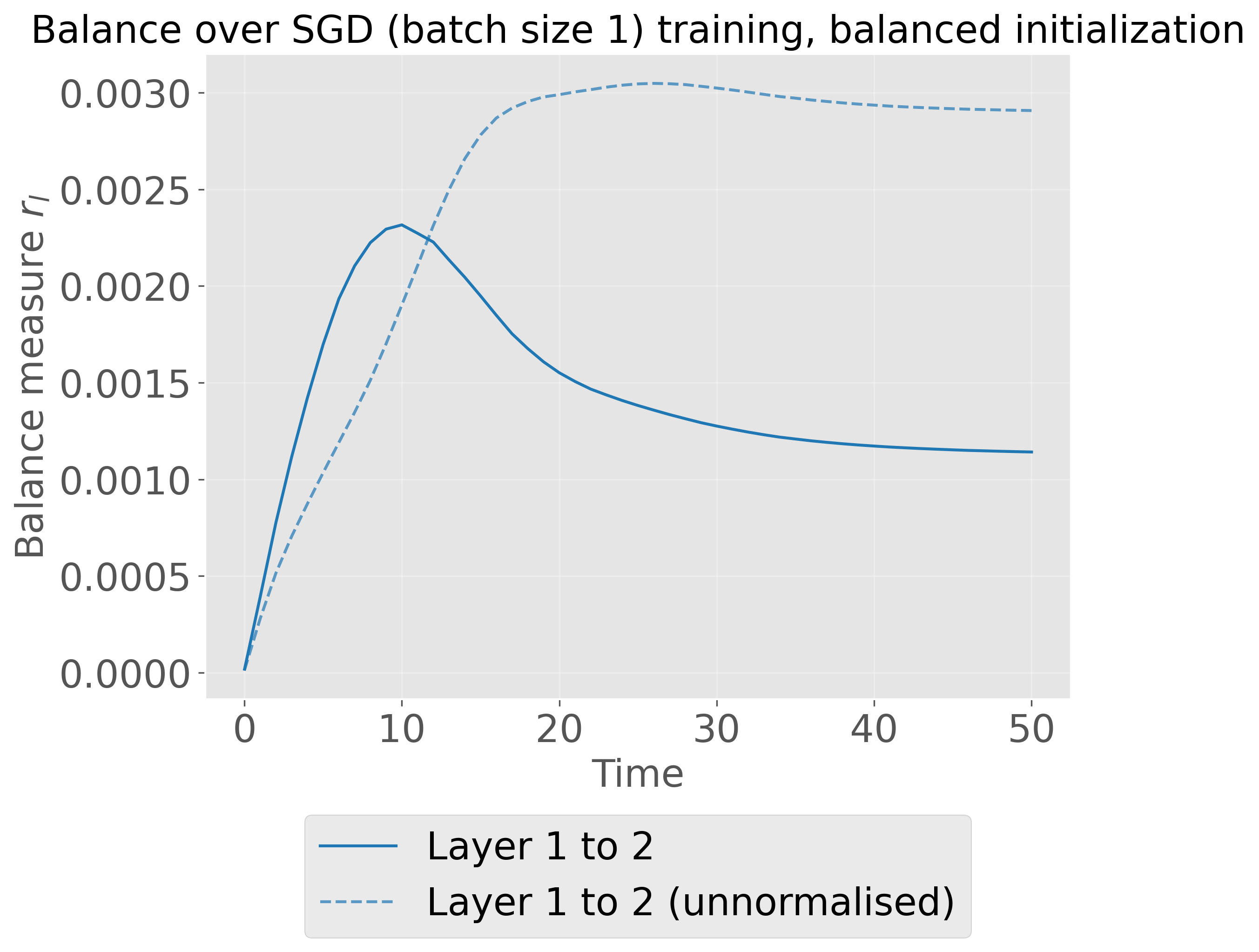}
        \caption{Balanced initialization, depth 2}
    \end{subfigure}
     \begin{subfigure}{0.48\textwidth}
        \includegraphics[width=\linewidth]{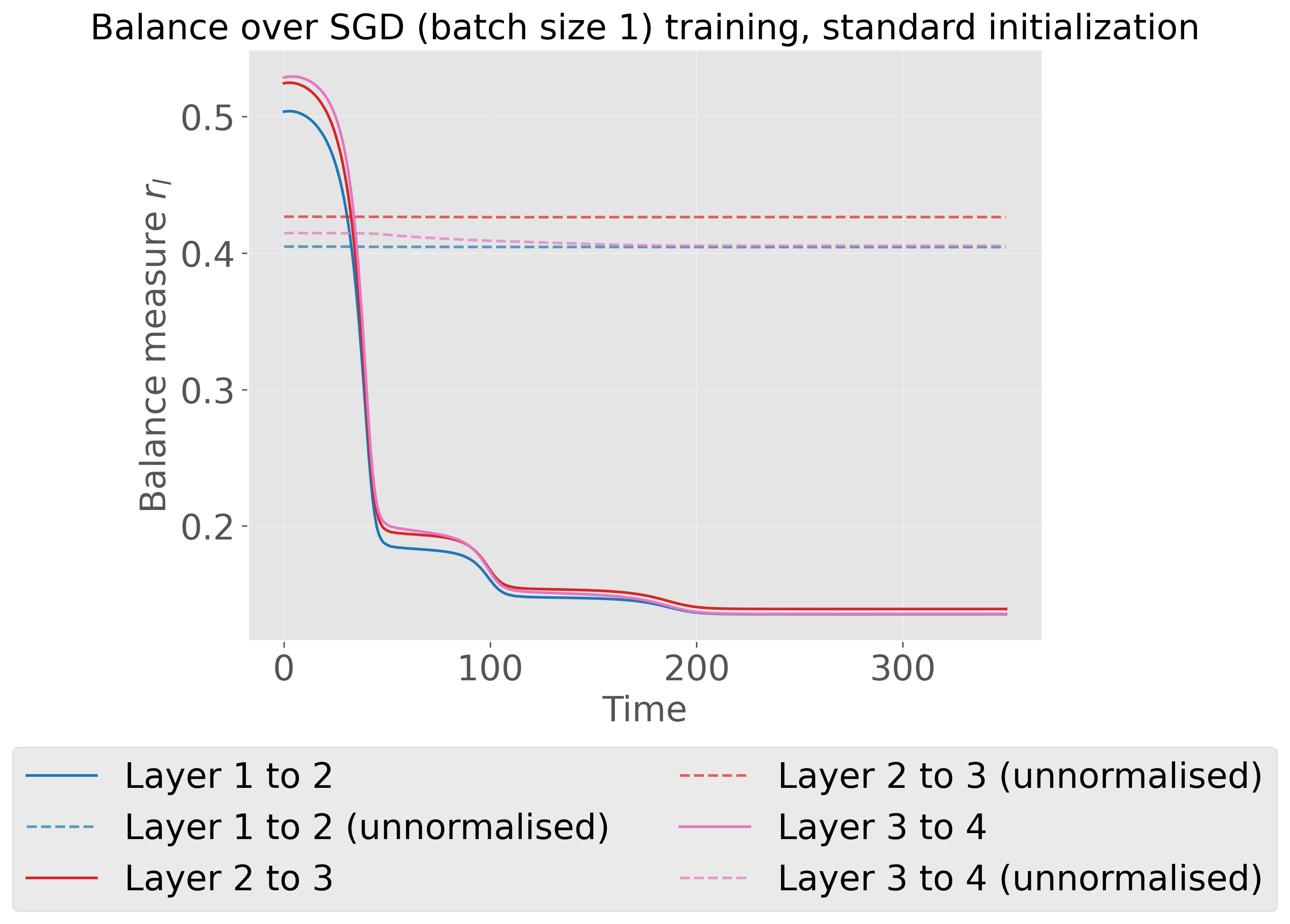}
        \caption{Standard initialization, depth 4}
    \end{subfigure}
    \begin{subfigure}{0.48\textwidth}
        \includegraphics[width=\linewidth]{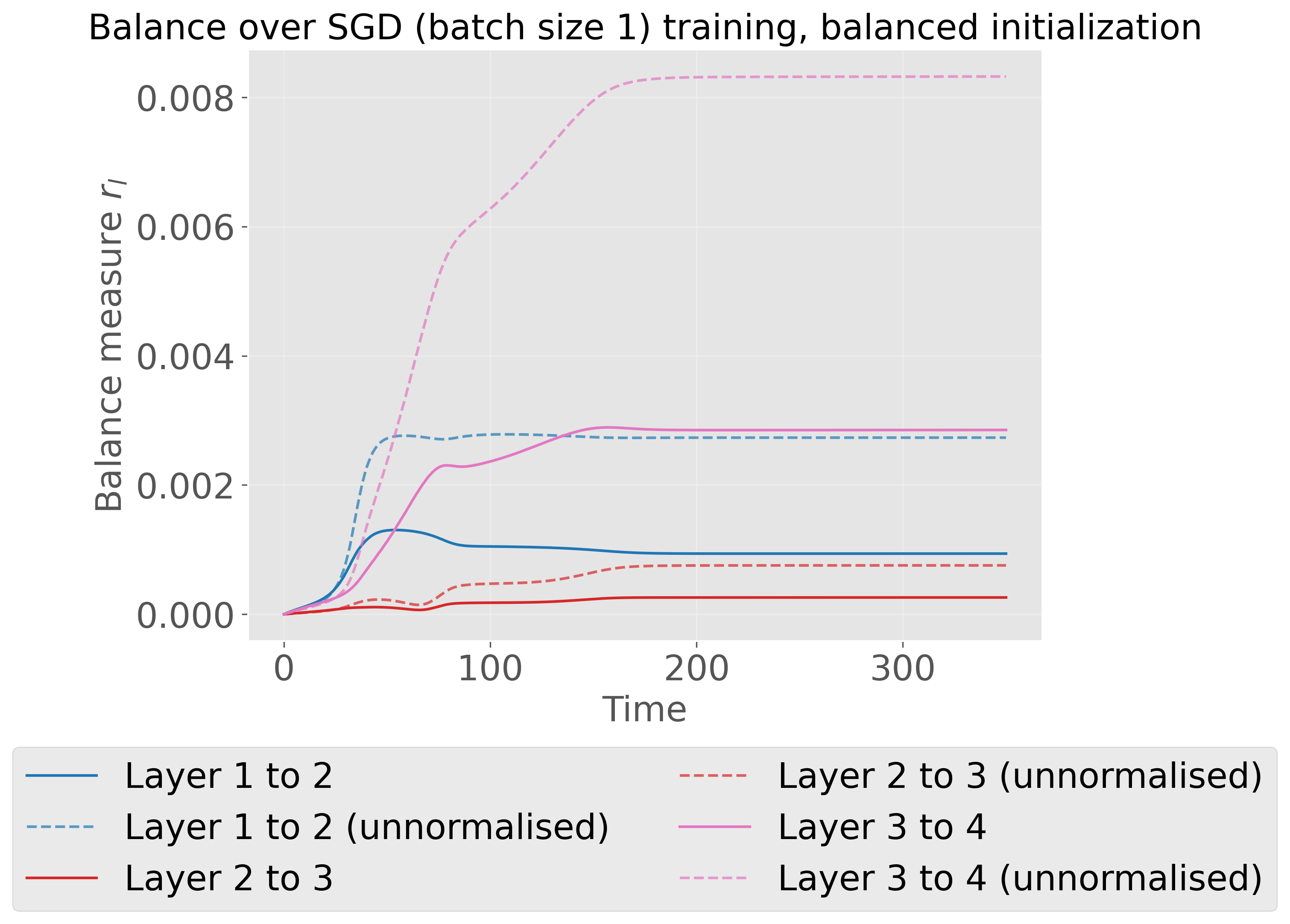}
        \caption{Balanced initialization, depth 4}
    \end{subfigure}
    \caption{\textbf{Measuring balance from balanced and unbalanced initialization}. (a) shows that from an unbalanced Gaussian initialization, there is not strong balance at the start of training, but as training continues balance increases. (b) When we enforce balance at initialization, it is approximately maintained, and for all of training the weights are significantly more balanced than at any point in the unbalanced-initialization trajectory. (c) and (d) show that this holds for deeper (depth-4) linear networks. The unnormalized lines are the numerator of the expression for $r_l$, and show little change.}
    \label{fig:balance_stats}
\end{figure}

\subsection{Balance}

We also test how well balance holds under non-balanced initializations. For this we track, for each non-final layer $l$,
\begin{equation}
    r_l := \frac{\lVert W_lW_l^\top - W_{l+1}^\top W_{l+1}\rVert_F} {\lVert W_lW_l^\top \rVert_F  + \lVert W_{l+1}^\top W_{l+1}\rVert_F} ,
    \label{eq:balance_measure}
\end{equation}
where small values of $r_l$ correspond to approximately balanced weights (normalized to be invariant to weight scaling).

\autoref{fig:balance_stats} shows that from non-balanced initializations, balance increases over SGD training of both a 2-layer and 4-layer linear network, but never reaches a level comparable to when balance is enforced at the start of training.

\begin{figure}
    \centering
    \begin{subfigure}{0.48\textwidth}
        \includegraphics[width=\linewidth]{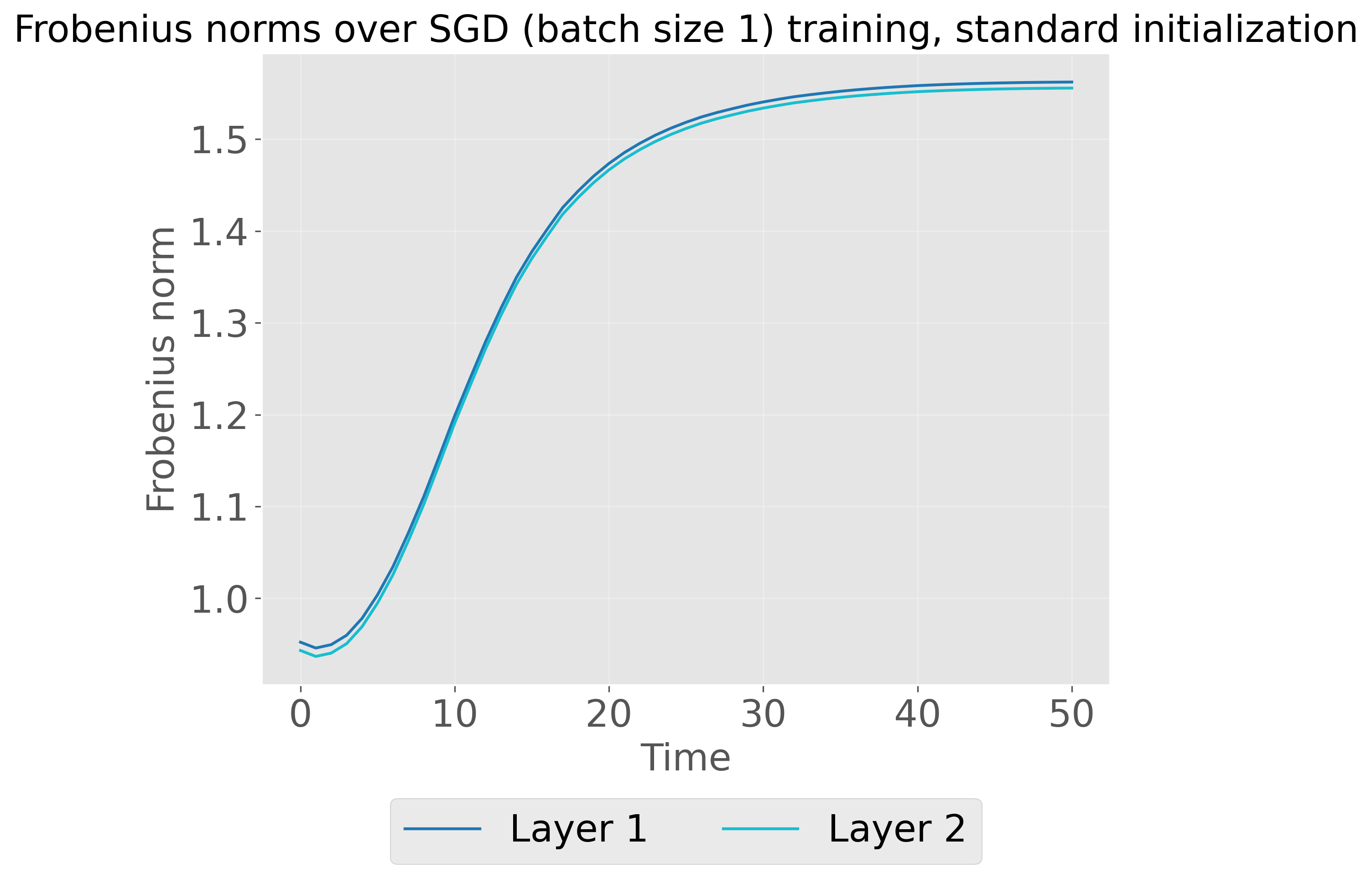}
        \caption{Standard initialization, depth 2}
    \end{subfigure}
    \begin{subfigure}{0.48\textwidth}
        \includegraphics[width=\linewidth]{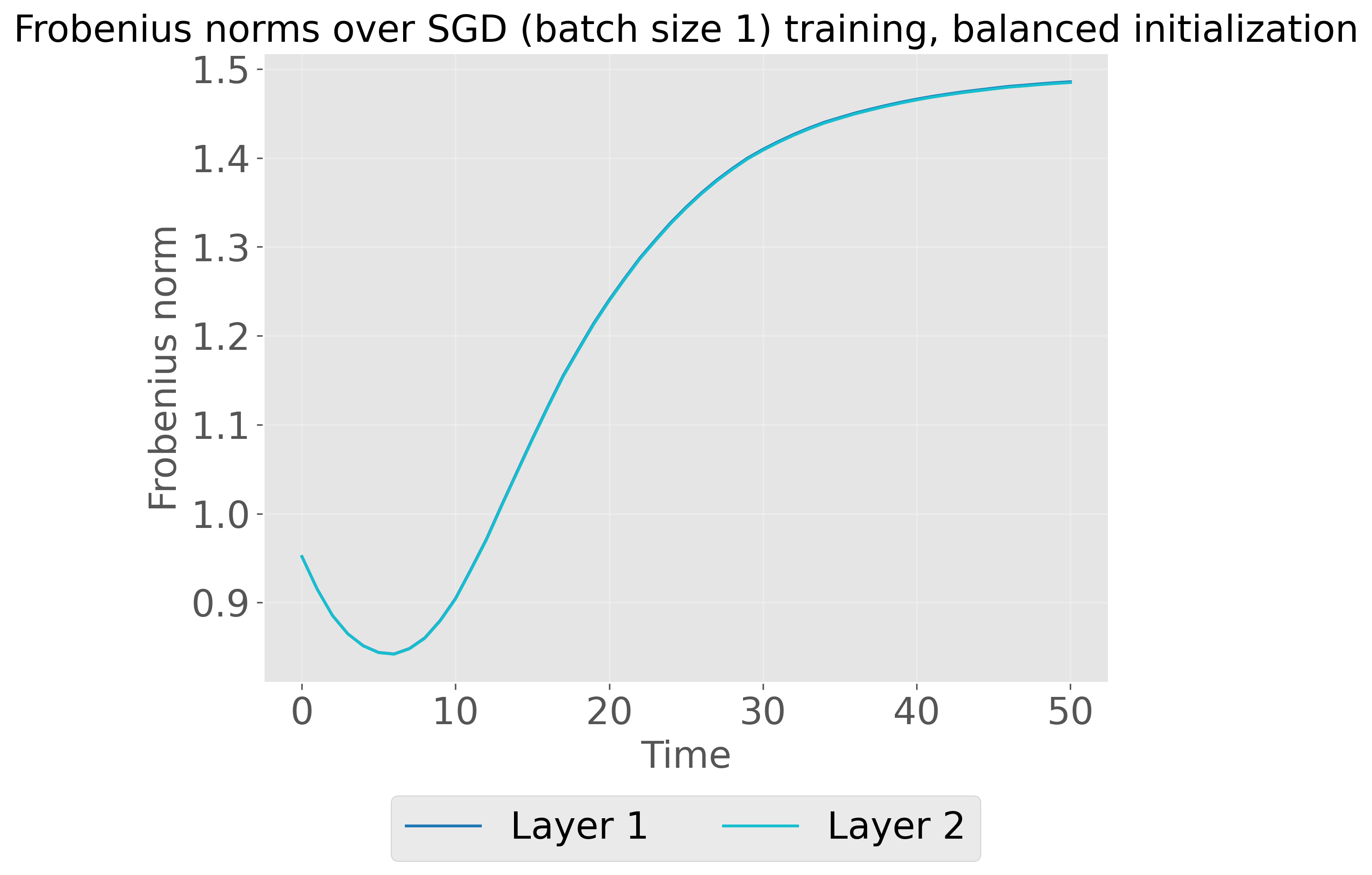}
        \caption{Balanced initialization, depth 2}
    \end{subfigure}
     \begin{subfigure}{0.48\textwidth}
        \includegraphics[width=\linewidth]{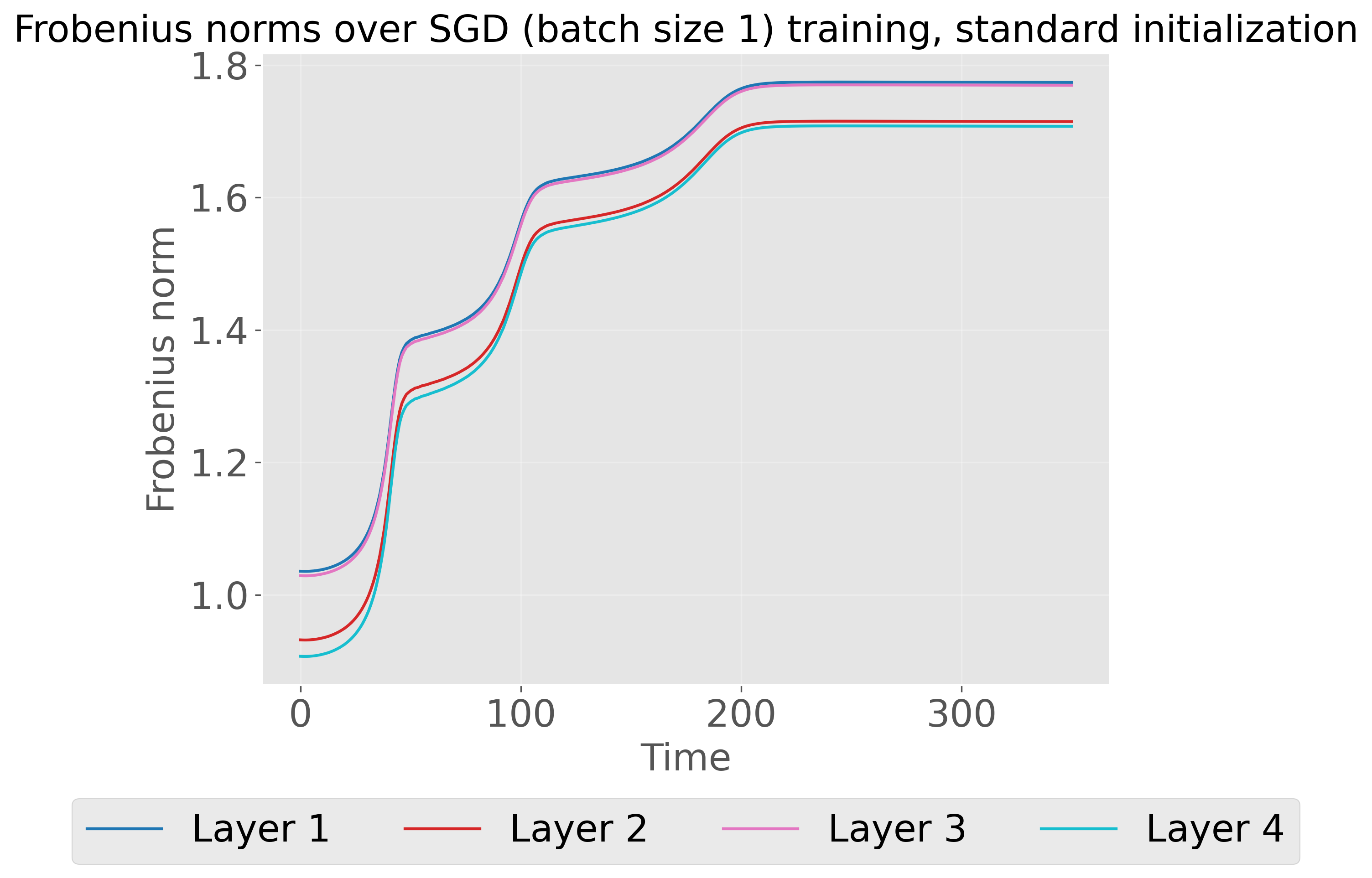}
        \caption{Standard initialization, depth 4}
    \end{subfigure}
    \begin{subfigure}{0.48\textwidth}
        \includegraphics[width=\linewidth]{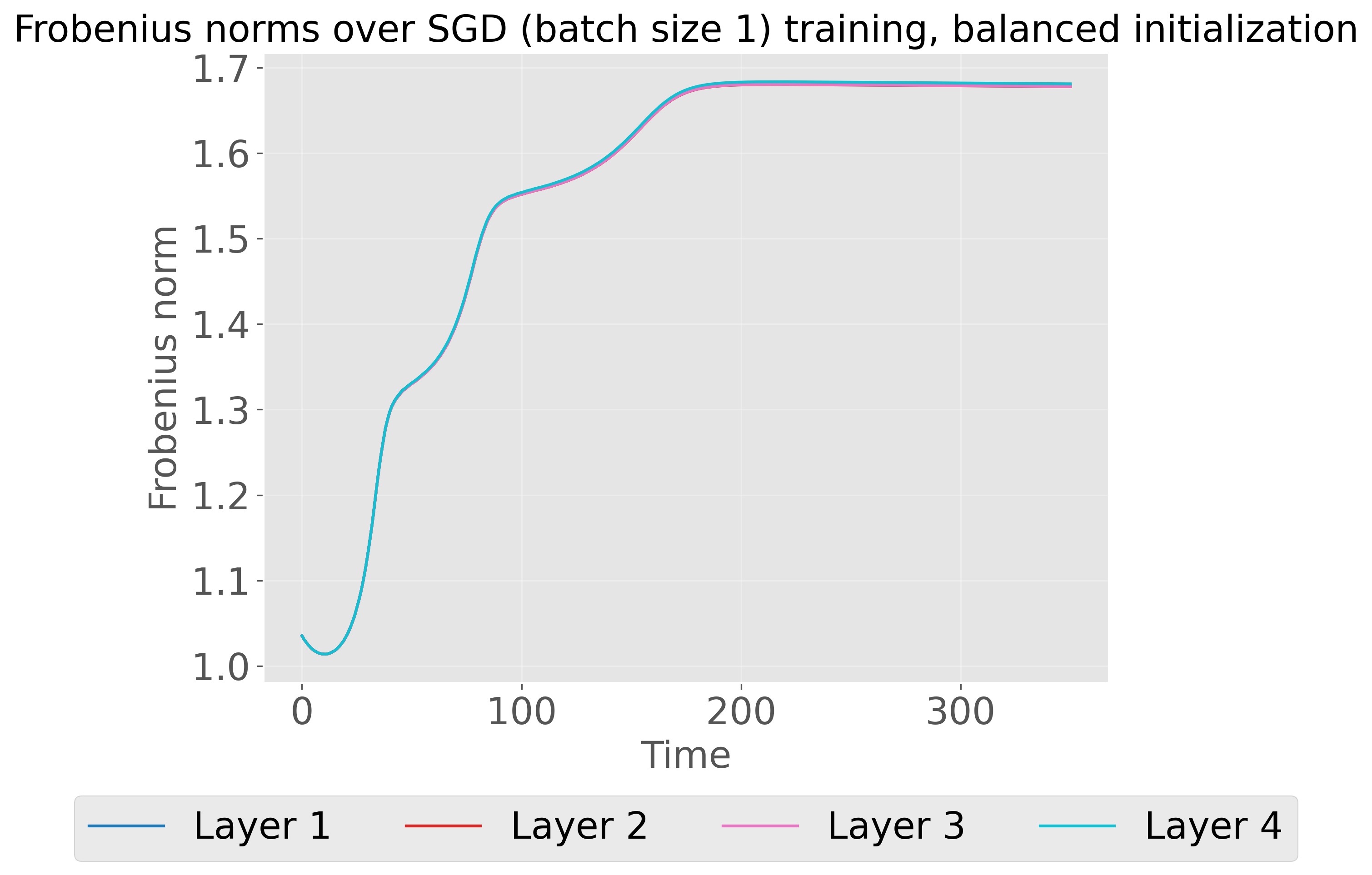}
        \caption{Balanced initialization, depth 4}
    \end{subfigure}
    
    \caption{\textbf{Frobenius norms of weight matrices over SGD training}. Frobenius norms of layers increase over training, accounting for most of the decrease in normalized balance measure $r_l$ from standard initializations.}
    \label{fig:frobenius_norms}
\end{figure}

To give a sense of the scale of the (im)balance, we also plot the unnormalized numerator of \autoref{eq:balance_measure} and also the Frobenius norm of each layer in \autoref{fig:frobenius_norms}. This reveals that the increase in balance from unbalanced initialization is mostly an increase relative to the increasing magnitude of the weights, and not an increase in absolute terms.

\section{Varying hyperparameters}
\label{app:varying-hyperparams}
 This appendix shows the effect of varying hyperparameters on our experimental results connecting mode learning with modewise diffusion.

 \subsection{Learning rate}

\begin{figure}
    \centering
    \includegraphics[width=\linewidth]{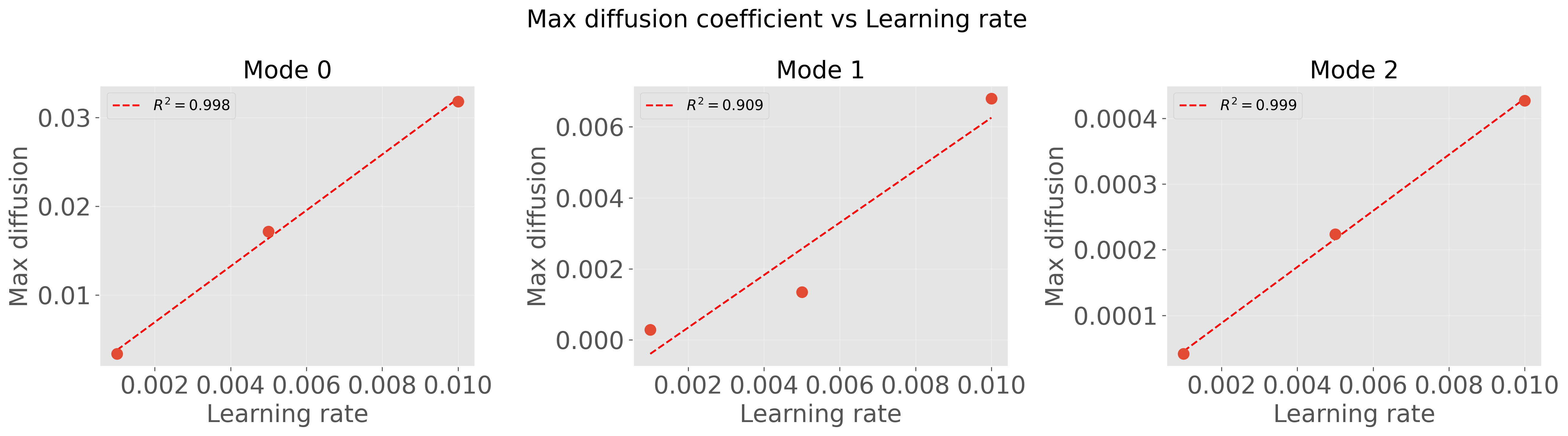}
    \caption{\textbf{Learning rate versus maximum diffusion for each mode.} We observe a linear relationship between learning rate and the diffusion along each mode.}
    \label{fig:app-vary-lr-diffusion}
\end{figure}

\autoref{fig:app-vary-lr-diffusion} shows that the maximum diffusion along modes scales approximately linearly with the learning rate, in agreement with the functional form derived in Proposition~\ref{prop:modewise-SDE-balanced}.

 \subsection{Batch size}

\begin{figure}
    \centering
    \includegraphics[width=\linewidth]{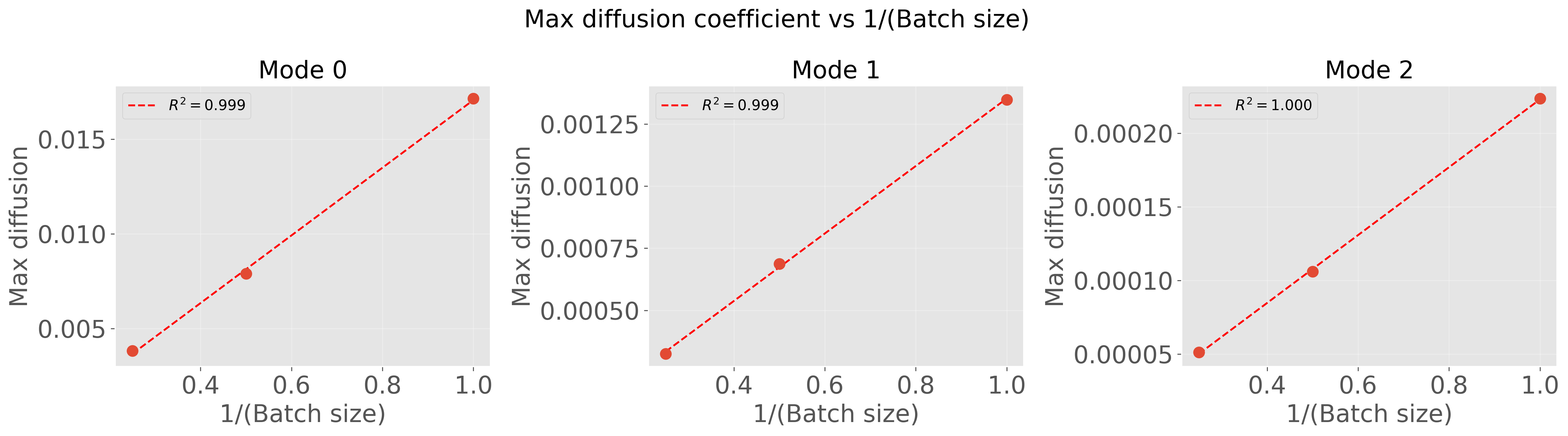}
    \caption{\textbf{Batch size versus maximum diffusion for each mode.} We observe a linear relationship between the reciprocal of batch size and the diffusion along each mode.}
    \label{fig:app-vary-bs-diffusion}
\end{figure}

 \autoref{fig:app-vary-bs-diffusion} shows that the maximum diffusion along modes scales approximately linearly with the reciprocal of the batch size, in agreement with the functional form derived in Proposition~\ref{prop:modewise-SDE-balanced}.

 \subsection{Finite dataset}

 This section shows that the results about the diffusion over time

 \subsection{DLN architecture}

\begin{figure}
    \centering
    \includegraphics[width=\linewidth]{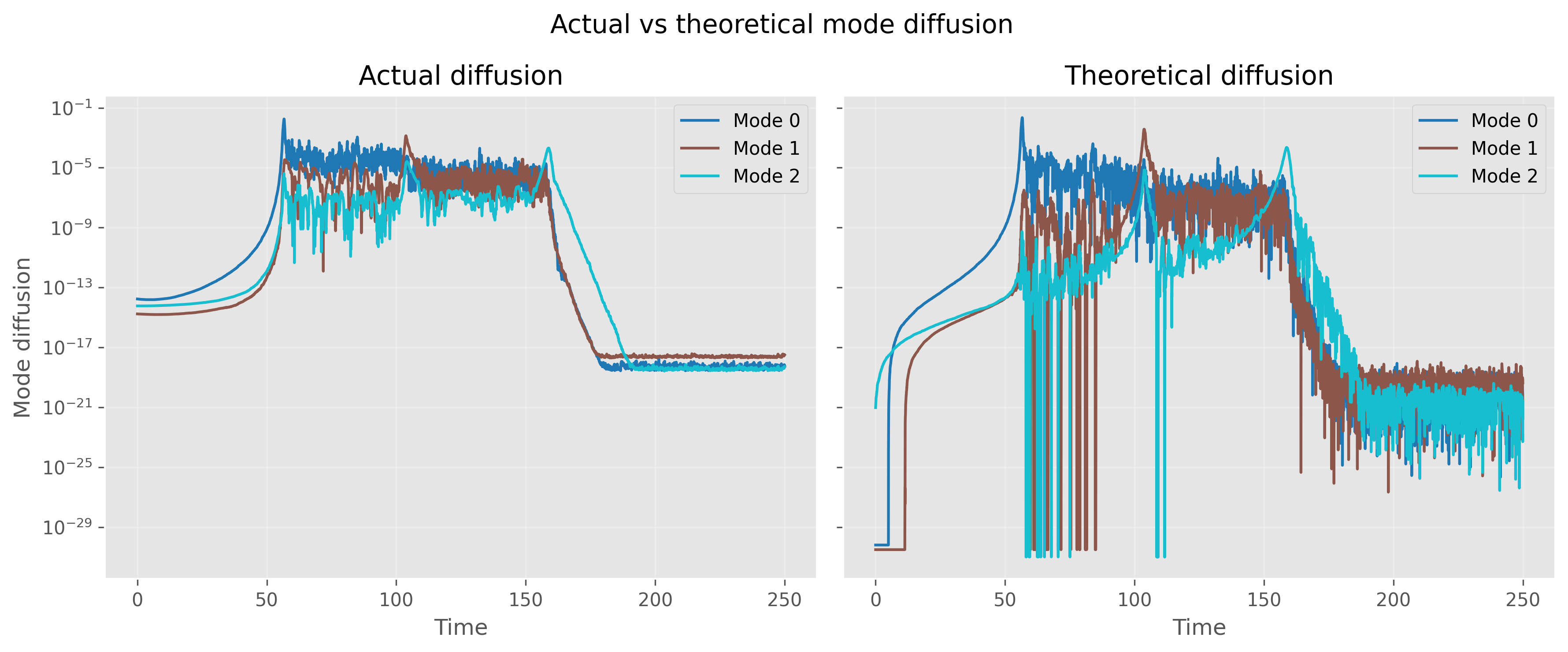}
    \caption{\textbf{Empirical versus theoretical modewise diffusion for a non-rectangular DLN.} The structure of the diffusion, including the location of the peaks and tending to zero once the mode is learned is maintained.}
    \label{fig:app-vary-width}
\end{figure}

So far, all experiments shown are with a rectangular DLN architecture (rectangular means that all the weight matrices are square). \autoref{fig:app-vary-width} shows that the formula in Proposition~\ref{prop:modewise-SDE-balanced} for the modewise diffusion is also similar to what is observed in a DLN with hidden layers of size 24, double the size of the input and output (12).

\section{Discretization error and end-of-training distribution}
\label{app:disc-error}

\begin{figure}
    \centering
    \includegraphics[width=0.5\linewidth]{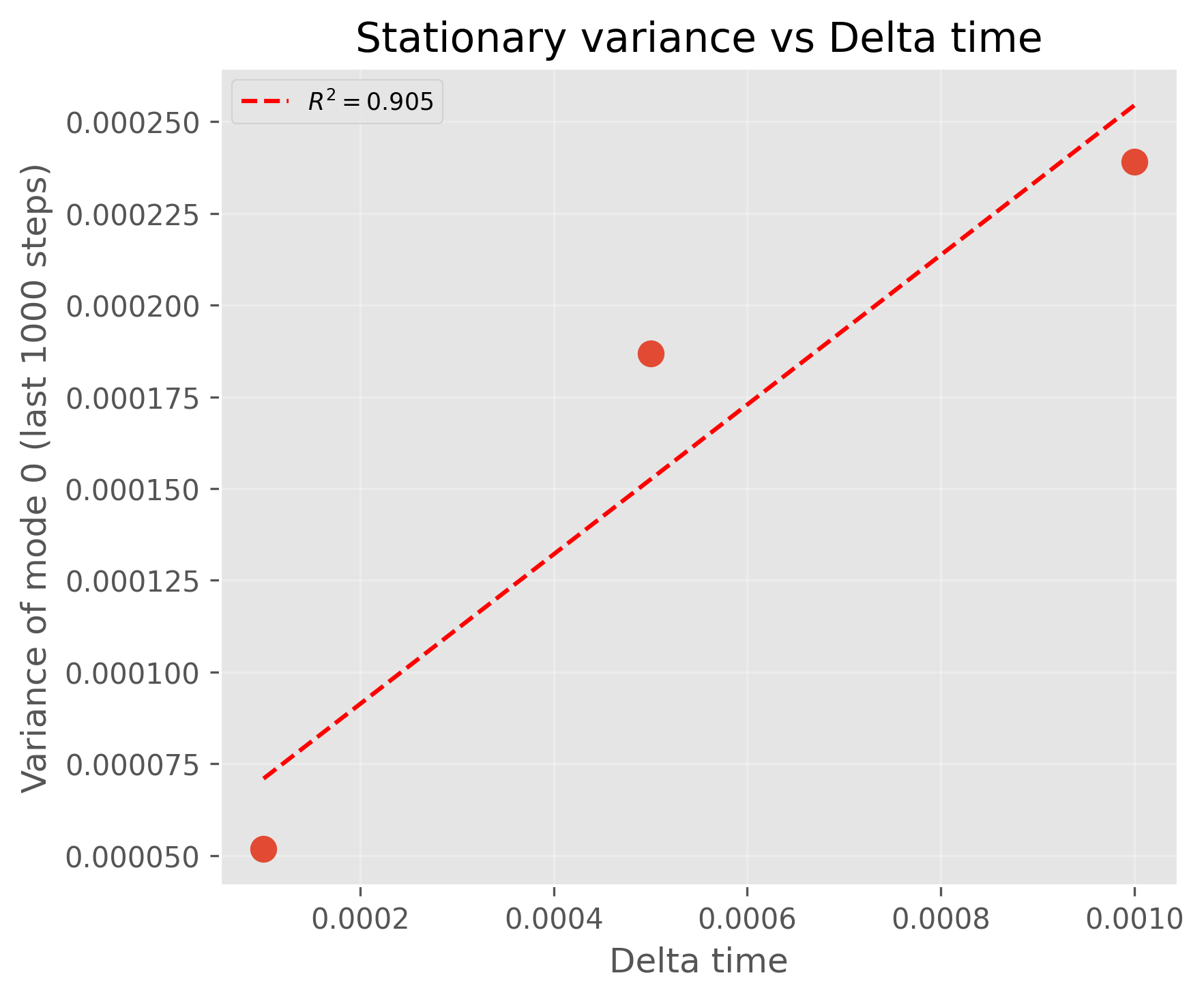}
    \caption{\textbf{Variance of the end-of-training distribution of the first mode versus simulation fineness.} We vary the fineness (i.e., the $\Delta t$ parameter) of the Euler-Maruyama simulation of the stochastic gradient flow SDE (\autoref{eq:sgd-sde}). For finer simulations, the variance reduces.}
    \label{fig:app-vary-dt}
\end{figure}

This appendix simulates the SDE in \autoref{eq:sgd-sde} with different values of $\Delta t$ for Euler-Maruyama. Smaller values of $\Delta t$ correspond to lower discretization error. At lower levels of discretization error, we observe that the end-of-training mode distribution has lower variance (\autoref{fig:app-vary-dt}) and is thus closer to the prediction of Proposition~\ref{prop:modewise-stationary}. This provides weak evidence in favor of our explanation for the mismatch between the prediction of Proposition~\ref{prop:modewise-stationary} and the experiments shown in \autoref{fig:distribution-convergence}.

\end{document}